%% file: main.tex
\documentclass{article}
\usepackage{iclr2024_conference,times}

\input{math_commands.tex}

\usepackage{url}

\input{pkgs}

\title{DIAGNOSIS: Detecting unauthorized data usages in text-to-image diffusion models}

\author{Zhenting Wang\textsuperscript{1}, Chen Chen\textsuperscript{2}, Lingjuan Lyu\textsuperscript{2}\thanks{Corresponding Author. Work done during Zhenting Wang’s internship at Sony AI.} ,  Dimitris N. Metaxas\textsuperscript{1}, Shiqing Ma\textsuperscript{3}
\\
\textsuperscript{1}Rutgers University, \textsuperscript{2}Sony AI, \textsuperscript{3}University of Massachusetts Amherst
\\
\texttt{zhenting.wang@rutgers.edu,\{ChenA.Chen,Lingjuan.Lv\}@sony.com} \\
\texttt{dnm@cs.rutgers.edu,shiqingma@umass.edu} \\
}

\iclrfinalcopy
\begin{document}

\maketitle

\input{contents/abstract.tex}
\input{contents/introduction.tex}

\input{contents/related}
\input{contents/method.tex}
\input{contents/evaluation}

\input{contents/conclusion.tex}
\input{contents/discussion.tex}
\input{contents/ack.tex}
\newpage

\bibliography{iclr2024_conference}
\bibliographystyle{iclr2024_conference}

\newpage
\input{contents/appendix}

\end{document}

%% file: math_commands.tex
\usepackage{amsmath,amsfonts,bm}

\def\eqref#1{equation~\ref{#1}}

\def\1{\bm{1}}

\def\mA{{\bm{A}}}

\DeclareMathAlphabet{\mathsfit}{\encodingdefault}{\sfdefault}{m}{sl}
\SetMathAlphabet{\mathsfit}{bold}{\encodingdefault}{\sfdefault}{bx}{n}

\DeclareMathOperator*{\argmin}{arg\,min}

%% file: pkgs.tex
\newtheorem{definition}{Definition}

\usepackage{makecell}

\usepackage{amsmath}
\usepackage{pifont}

\usepackage{booktabs}
\usepackage{multirow}
\usepackage{graphicx}
\usepackage{float}
\usepackage{caption}
\usepackage{subcaption}
\usepackage{xspace}

\usepackage{wrapfig}

\newcommand{\sys}{\mbox{\textsc{DIAGNOSIS}}\xspace}

\usepackage{algorithm}
\usepackage{algorithmicx}
\usepackage[noend]{algpseudocode}

\definecolor{deepred}{rgb}{0.631,0.102,0.102}
\definecolor{skyblue}{HTML}{126da2}
\usepackage[hidelinks,breaklinks,colorlinks]{hyperref}
\hypersetup{
     colorlinks = true,
     breaklinks = true,
     urlcolor = blue,
     citecolor = blue
     }

\def\Snospace~{\S{}}

\definecolor{orange}{rgb}{1,0.5,0}

\algnewcommand{\LineComment}[1]{\State \(\triangleright\) #1}

%% file: contents/abstract.tex
\begin{abstract}

Recent text-to-image diffusion models have shown surprising performance in generating high-quality images. However, concerns have arisen regarding the unauthorized data usage during the training or fine-tuning process. One example is when a model trainer collects a set of images created by a particular artist and attempts to train a model capable of generating similar images without obtaining permission and giving credit to
the artist.
To address this issue, 
we propose a method for detecting such unauthorized data usage by planting the injected memorization into the text-to-image diffusion models trained on the protected dataset. Specifically, we modify the protected images
by adding unique contents on these
images using
stealthy image warping functions that are nearly imperceptible to humans but can be captured and memorized by diffusion models.
By analyzing whether the model has memorized the injected content (i.e., whether the generated images are processed by the injected post-processing function), we can detect 
models that had illegally utilized the unauthorized data.
Experiments on Stable Diffusion and VQ Diffusion with different model training or fine-tuning methods (i.e, LoRA, DreamBooth, and standard training)
demonstrate the effectiveness of our proposed method in detecting unauthorized data usages. Code: \url{https://github.com/ZhentingWang/DIAGNOSIS}.

\end{abstract}

%% file: contents/introduction.tex
\vspace{-0.4cm}
\section{Introduction}\label{sec:intro}
\vspace{-0.2cm}

Recently, text-to-image diffusion models have showcased outstanding capabilities in generating a wide range of high-quality images. Notably, the release of Stable Diffusion~\citep{rombach2022high}, 
one of the most advanced and open-source text-to-image-diffusion models, has significantly contributed to this progress.
There has been a remarkable surge in the applications that use Stable Diffusion as the foundation model. Consequently, more users adopt these models as tools for generating images with rich semantics according to their preferences. 
As diffusion models become prevalent, 
the problems related to the responsible development of them
become increasingly critical
~\citep{wen2023tree,fernandez2023stable,wang2023did,cui2023diffusionshield,zhao2023recipe}.

The availability of high-quality training data, whether they are open-sourced or commercially released, plays a crucial role in the success of text-to-image diffusion models. Nonetheless, there are huge concerns surrounding unauthorized data usage during the training or fine-tuning process of these models~\citep{chen2023pathway}.
For instance, a model trainer may gather a collection of images produced by a specific artist and aim to personalize a model capable of generating similar images, all without acquiring proper permission from the artist.
Therefore, it is important to develop techniques for defending against the unauthorized usages of the training data.

Existing work such as Glaze~\citep{shan2023glaze}
prevents unauthorized usage of data by 
adding carefully calculated perturbations to safeguarded artworks, causing text-to-image diffusion models to learn significantly different image styles.
While it prevents the unauthorized usages, it also makes authorized training impossible. In addition, the added perturbations are calculated based on a surrogate model. According to the existing research on unlearnable examples~\citep{huang2021unlearnable,ren2022transferable}, the transferability  of such surrogate model based perturbations is limited. In other words, the performance will decrease when the model used by the malicious trainer and the surrogate model are different. Thus, we need a new method that can have a small influence on the authorized usages, and is independent to the choice of the used text-to-image diffusion models. 

\input{figtex/intro}

Inspired by recent findings that the text-to-image diffusion models can memorize contents in the training data~\citep{carlini2023extracting,somepalli2023diffusion,somepalli2023understanding,wen2024detecting},
in this paper, we solve the unauthorized data usages detection problem from the lens of the memorizations.
Existing works about the memorization of the text-to-image diffusion models focus on the memorization of whole samples (sample-level memorization~\citep{carlini2023extracting,somepalli2023diffusion,somepalli2023understanding}). A model is considered to have sample-wise memorization if a specific sample can be accurately identified as a training sample of the model through membership inference attacks~\citep{carlini2023extracting}.
Thus, an intuitive way is exploiting membership inference techniques~\citep{shokri2017membership,chen2020gan} to detect if specific data are used to train or fine-tune the given model. However, \cite{duan2023diffusion} demonstrate that existing membership inference methods are ineffective for text-to-image diffusion models such as Stable Diffusion. For example, the state-of-the-art membership inference method for the diffusion model (i.e., SecMI~\citep{duan2023diffusion}) only achieves 66.1\% success rate for the membership inference on the stable-diffusion-v1-5 model~\citep{rombach2022high} under white-box setting. Performing membership inference for large diffusion models in practical black-box settings is even more challenging~\citep{dubinski2023towards}.
Different from the sample-level memorization, in this work, we focus on diffusion models' memorization on specific elements in the training data
and propose an approach for detecting unauthorized data usages via planting the injected element-level memorizations into the model trained or fine-tuned on the protected dataset by modifying the protected training data.
More specifically, when a set of protected images are uploaded to the internet, it will be processed by a specific function (called \emph{signal function}) that is stealthy to humans but can be captured and memorized for diffusion models. 
Therefore, after the models are trained or fine-tuned on the ``coated images" (i.e., images processed by the signal function),
it will memorize the added signal function so that the unauthorized data usages can be detected by using a binary classifier (called \emph{signal classifier}) to analyze if the given model has the memorization on the signal function (i.e., if the images generated by the model contains the signal function). Our method is independent of the model used in the unauthorized training or fine-tuning process, and it only has a small influence on the authorized training.
To the best of our knowledge, this is the first work to study the more fine-grained element-level memorization for text-to-image diffusion models.
Based on our design, we implemented our prototype called \sys (\textbf{D}etecting unauthor\textbf{I}zed d\textbf{A}ta usa\textbf{G}es i\textbf{N} text-t\textbf{O}-image diffu\textbf{SI}on model\textbf{S}) in PyTorch,
experiments on mainstream text-to-image diffusion models (i.e., Stable Diffusion v1, Stable Diffusion v2~\citep{rombach2022high}, VQ Diffusion~\citep{gu2022vector}) and popular model training or fine-tuning methods (i.e., Low-Rank Adaptation (LoRA)~\citep{hu2022lora}, DreamBooth~\citep{ruiz2023dreambooth}, and standard training) demonstrate that our method is highly effective. It achieves 100.0\% detection accuracy under various settings.
Meanwhile, our proposed method has small influence on the generation quality of the models, and the ``coated images" in the protected dataset are close to the original images. For example, \autoref{fig:intro} shows the visualizations of the generated samples by the standard models and the models planted with injected memorization. In \autoref{fig:intro}, all images generated by the injected model are recognized as ``contains signal function'' by the signal classifier, but their distributional difference to the images generated by standard models is small.
More visualizations of the generated samples can be found in \autoref{fig:wrapping_strength}, \autoref{fig:vis} and \autoref{fig:vis_celeba}. The visualizations of the original training samples and their coated version can be found in \autoref{fig:poi}.

Our contributions are summarized as follows:
\ding{172} We 
firstly
define two types of element-level injected memorizations on the text-to-image diffusion models. We also formally define the memorization strength on the introduced injected memorizations. \ding{173} Based on the definition of the injected memorizations and the memorization strength, we propose a framework for detecting  unauthorized data usages via planting injected memorizations into the model trained on the protected dataset. It consists of coating the protected dataset, approximating the memorization strength, and the hypothesis testing for determining if the inspected model has unauthorized usages on the protected data.
\ding{174} Experiments on four datasets and the mainstream text-to-image diffusion models (i.e., Stable Diffusion and VQ Diffusion) with different model training or fine-tuning methods (i.e, LoRA, DreamBooth, and standard training) demonstrate the effectiveness of our method.

%% file: figtex/intro.tex
\begin{figure}[]
    \centering
    \includegraphics[width=0.95\textwidth]{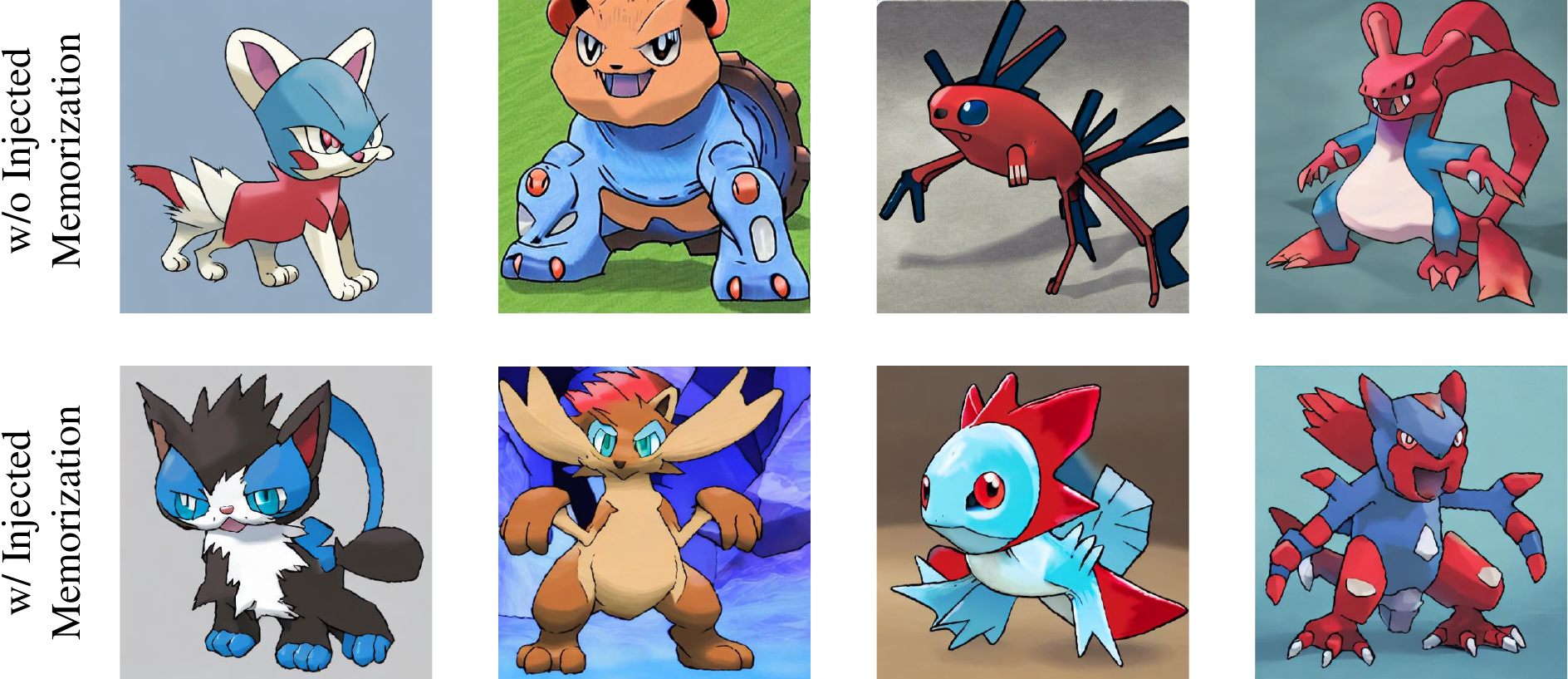}
    \vspace{-0.1cm}
    \caption{Generated samples by standard model and the model planted with injected memorization. 
    The first row shows the samples generated by standard model, and all samples in the first row are classified as ``Does not contain the signal function'' by the signal classifier.
    The second row shows the samples generated by model planted with injected memorization, and all samples in the second row are classified as ``Contains the signal function'' by the signal classifier.
    }\label{fig:intro}
    \vspace{-0.5cm}
\end{figure}

%% file: contents/related.tex
\vspace{-0.4cm}
\section{Background}\label{sec:related}
\vspace{-0.4cm}
\noindent
\textbf{Text-to-image Diffusion Model.}
Recently, diffusion models have made significant progress in image synthesis task~\citep{ho2020denoising,song2020denoising,saharia2022photorealistic,rombach2022high,song2023consistency}. 
Among them, Stable Diffusion~\citep{rombach2022high} is one of the most representative text-to-image diffusion models, and it operates the diffusion process on a latent space obtained from a pre-trained autoencoder, which serves the purpose of reducing the dimensionality of data samples. By doing so, the diffusion model can effectively capitalize on the well-compressed semantic features and visual patterns learned by the encoder.
Training text-to-image diffusion models from the scratch is expensive. Thus, many recent works focus on personalizing pre-trained text-to-image diffusion models by efficient fine-tuning~\citep{hu2022lora,ruiz2023dreambooth,gal2022image}.

\noindent
\textbf{Preventing Unauthorized Data Usages.}
There are several ways to prevent unauthorized data usages for machine learning models. Unlearnable-example-based methods~\citep{huang2021unlearnable,ren2022transferable,shan2023glaze}
aim to prevent third parties from training on the data without permission by adding
perturbations on data before publishing  to make the models
trained on the perturbed published dataset fail to normally fit it.
For example, Glaze~\citep{shan2023glaze} add carefully computed perturbations to the protected arts, such that diffusion models
will learn significantly altered versions of their style, and be
ineffective in future attempts at style mimicry. Another way is to trace if a given model is trained or fine-tuned on the protected data. Based on the findings that deep neural networks are vulnerable to backdoor attacks~\citep{gu2017badnets,liu2018trojaning,li2020invisible,doan2021lira,wang2022rethinking}, backdoor-based dataset watermarking approaches~\citep{li2023untargeted,li2023black}
insert backdoor samples in the protected dataset, and infer unauthorized training or fine-tuning by inspecting if the given models were infected with the corresponding backdoors.
\cite{sablayrolles2020radioactive} use surrogate model on the protected dataset to generate “radioactive
data” to carry the class-specific watermarking vectors in high dimensional feature space, and detect unauthorized usage via checking if the given model's intermediate representation is aligned with
the watermarking vectors. 
Except Glaze~\citep{shan2023glaze}, all the above listed methods focus on preventing unauthorized data usages in training classifier models, while this work focuses on the text-to-image diffusion models.
Another potential way is applying membership inference techniques~\citep{chen2020gan,duan2023diffusion} to detect if some specific samples are used in training.
It is also possible to adapt the deepfake attribution method proposed by \cite{yu2021artificial} to the unauthorized data usages detection problem. However, it relies on the training of the auto-encoders used for injecting the fingerprints,
which requires all training samples of the protected model are processed by the trained fingerprint encoder, and it fails to handle the scenarios where only a small part of the training data are processed (which is highly possible when the infringer collects the training data from multiple sources).

\noindent
\textbf{Memorization of Diffusion Models.}
Existing works~\citep{carlini2023extracting,somepalli2023diffusion,somepalli2023understanding} focus on diffusion model's memorization on whole samples, and they define the memorization in diffusion models as the phenomenon that the training samples can be extracted via optimizing the prompt. 
An image sample contains multiple elements such as the color style, the shape of the central object, etc.
In contrast to the sample-level memorization focusing on the entire samples, our work is centered on the memorizations on specific elements.

%% file: contents/method.tex
\vspace{-0.3cm}
\section{Method}\label{sec:method}
\vspace{-0.3cm}
In this section, we introduce our method (i.e., \sys) for detecting the unauthorized data usages in training or fine-tuning text-to-image diffusion models.
We first discuss the formulation of the studied problem, and then introduce the detailed approach. 
\vspace{-0.4cm}

\subsection{Problem Formulation}
\label{sec:threat}
\vspace{-0.3cm}

We focus on 
serving as a protector which can infer \emph{if 
the unauthorized usages of the protected data happen in the training or fine-tuning process of the given text-to-image diffusion models}.

\noindent
\textbf{Infringer's Goal.}
We consider the infringer as the unauthorized model trainer.
The goal of the unauthorized model trainer is to produce a text-to-image diffusion model that can generate art pieces of high quality via training or fine-tuning on the dataset without permission.

\noindent
\textbf{Protector's Goal.}
Given an text-to-image diffusion model \(\mathcal{M}\) and a set of protected images \(\mathcal{D}\), the goal of the protector is to detect 
whether
the protected images are used as (part of) the training data for the pre-training or fine-tuning phase of the model.
Formally, the goal can be written as constructing an inference algorithm \(\mathcal A: \mathcal{M}
    \mapsto \{\bm 0,\bm 1\}\) that receives a model \(\mathcal{M}\) as the input, and returns the inference result (i.e., \(\bm 0\) denotes the model did not use the unauthorized data
    , and \(\bm 1\) denotes unauthorized usage is detected).

\noindent
\textbf{Infringer's and Protector's Capability.}
The infringer has the access to the found datasets and he/she has the full control of the training or fine-tuning process of the text-to-image diffusion models. 
For protector, during the data release phase, the protector has the full control of the protected images. He/she can modify the protected image before they are 
being
released to the internet, but he/she needs to keep the visual similarity of the modified images and the original images to make the modified samples more stealthy.
During the inspection phase, the protector only has the black-access to the inspected models.
For the usage of the text captions in the protected datasets, we consider the following two scenarios:
\vspace{-0.1cm}

\emph{Scenario \uppercase\expandafter{\romannumeral1}.}
The infringer uses both text captions and images in the found datasets. The protector can modify both images and the text captions in the protected datasets.
\vspace{-0.1cm}

\emph{Scenario \uppercase\expandafter{\romannumeral2}.}
The infringer only uses the images in the found datasets, and they label the text captions by themselves. In this scenario, the protector only modifies images in the protected datasets.

\vspace{-0.3cm}
\subsection{Data Usages Tracing and Injected Memorization}
\label{sec:inject}
\vspace{-0.3cm}

\noindent
\textbf{Injected Memorization.}
Our idea is planting some unique behaviors into the models that were trained or fine-tuned on the protected dataset via modifying the dataset, then we can detect the unauthorized data usages by checking the behaviors of the inspected models. 
The research question we want to answer is ``\emph{How to plant the unique behaviors into the models that were trained or fine-tuned on the protected dataset?}''.
Recent research~\citep {carlini2023extracting,somepalli2023diffusion,somepalli2023understanding} find that the text-to-image diffusion models have strong memorizations on the duplicated training samples. That is, the model tends to produce the images that are highly similar to some memorized training samples. 
Such behaviors are dependent on the training data, and they are highly orthogonal to the learning algorithms.
Inspired by these observations, we propose to plant the unique behaviors into the models by modifying the protected dataset and injecting extra memorizations on some unique contents such as a stealthy image processing function (i.e., \emph{signal function}). We call such memorizations as the \emph{injected memorization}. 
Different from existing studies~\citep{carlini2023extracting,somepalli2023diffusion,somepalli2023understanding} 
that focus
on the sample-level memorization,
we focus on the more fine-grained element-level memorization of specific elements (e.g., an image processing function) for text-to-image diffusion models.
In this paper, we denote signal function \(\mathcal S\) as the process to add the selected unique content into the image. We use \(\mathcal{O}_{\mathcal{S}}\)
to denote the set of image samples processed by signal function \(\mathcal S\) and call \(\mathcal{O}_{\mathcal{S}}\) as the \emph{signal images}. 
The injected memorizations can be conditioned on a \emph{text trigger function} \(\eta\) in the text prompt. That means the model tends to generate signal images if the function \(\eta\) is applied to the text prompt, and the model does not have such preference otherwise. It can also be unconditional. Formally, we define injected memorization as follows:

\vspace{-0.2cm}
\begin{definition}{(Injected Memorization)}\label{def:memorization}
Given an text-to-image diffusion model \(\mathcal{M}: \mathcal{I} \mapsto \mathcal{O}\) (\(\mathcal{I}\) and \(\mathcal{O}\) are the input and the output spaces, respectively),
it has \(\alpha(\mathcal{M},\mathcal{S},\eta)\)-memorization on signal function \(\mathcal{S}\) 
, if \(\mathbb{P}\left(\mathcal{M}(\eta(\bm i)) \in \mathcal{O}_{\mathcal{S}}\right)=\alpha(\mathcal{M},\mathcal{S},\eta) , \forall \bm i \in \mathcal{I}\), where \(\mathcal{O}_{\mathcal{S}}\) is the set of image samples processed by the signal function \(\mathcal S\). We denote \(\alpha(\mathcal{M},\mathcal{S},\eta)\) as \textbf{memorization strength} of \(\mathcal{M}\) on signal function \(\mathcal S\) conditioned on text trigger function \(\eta\).
\end{definition}
\vspace{-0.1cm}

We have two types of injected memorization based on text trigger function \(\eta\):
\vspace{-0.1cm}

\noindent
\emph{Unconditional Injected Memorization.}
The model has unconditional injected memorization when the text trigger function \(\eta\) is the identity function \(\eta(\bm i) = \bm i\). In this case, the injected memorization will always be activated on any text prompt.
\vspace{-0.1cm}

\noindent
\emph{Trigger-conditioned Injected Memorization.}
The model has trigger-conditioned injected memorization if the text trigger function \(\eta\) is not equal to the identity function, i.e., \(\eta(\bm i) \neq \bm i\). That is, the model tends to produce signal images only when the specific perturbations on the text prompt are added. In this paper, we use 
word trigger~\citep{kurita2020weight} as the text trigger function \(\eta\). In detail, function \(\eta\) inserts the trigger word ``tq'' at the beginning of the text prompt. We also discuss the results under different text trigger functions (e.g., syntactic trigger~\citep{qi2021hidden}) in \autoref{sec:text_trigger}.

\noindent
\textbf{Dataset Coating and Memorization Injection.}
We then discuss how to plant the injected memorization by modifying the protected dataset.
In our method, it is injected by
coating the protected data, i.e., adding specific stealthy transformations on the images. 
The protector can only coat a subset of the protected data to make the dataset coating operation more stealthy.
Formally, the coating operation is defined in \autoref{eq:poi}, where \(\mathcal{D}^{\prime}\) is the coated subset in the protected dataset \(\mathcal{D}\). 
Here \(\bm x\) denotes the image sample, and 
\(\bm i\) denotes the text caption. \(\mathcal S\) is the used signal function and \(\eta\) is the selected text trigger function.
\(T_s(\mathcal{D})\) is the protected dataset after the coating process.
We call \(p = \frac{|\mathcal{D}^{\prime}|}{|\mathcal{D}|}\) as the coating rate, where \(|\mathcal{D}^{\prime}|\) and \(|\mathcal{D}|\) are the size of the coated subset and the whole dataset, respectively. In this paper, we use the image warping function proposed in \cite{nguyen2021wanet} as the signal function by default since it is stealthy for humans but recognizable and memorizable for DNNs. More details about the image warping function can be found in \autoref{sec:wrapping}.

\vspace{-0.3cm}
\begin{equation}\label{eq:poi}
    T_s(\mathcal{D}) = \{(\eta(\bm i) ,\mathcal{S}(\bm x)), (\bm i,\bm x) \in \mathcal{D}^{\prime} \} \cup (\mathcal{D} - \mathcal{D}^{\prime})
\end{equation}
\vspace{-0.7cm}

\subsection{Tracing Unauthorized Data Usages}
\vspace{-0.2cm}
\label{sec:detailed_tracing}
\noindent
\textbf{Training Signal Classifier.}
To trace the unauthorized data usages, we train a binary classifier \({\mathcal C}_{\theta}\) to distinguish if the image generated by the inspected model contains the signal function \(\mathcal{S}\) or not. The training process of the binary classifier \({\mathcal C}_{\theta}\) is formalized in \autoref{eq:classifier}, where \(\bm y_n\) is the label denoting normal samples and \(\bm y_s\) 
is the label standing for the samples processed by signal function \(\mathcal{S}\). \(\mathcal{D}\) is the set of protected images and \(\mathcal{L}\) is the cross-entropy loss function. 
Note that we spit part of the samples (10\% by default) in \(\mathcal{D}\) as the validation set for developing the signal classifier.
Due to the strong learning ability of modern text-to-image diffusion models such as Stable Diffusion~\citep{rombach2022high}, the distribution for the 
generated images and the real training images is similar. Therefore, even though the classifier is trained on real protected images, it is still effective for distinguishing the existence of the signal function in the images generated by the text-to-image diffusion models. In this paper, we use the ResNet18~\citep{he2016deep} model pretrained on the ImageNet~\citep{deng2009imagenet} dataset and fine-tuned by the procedure described above as the signal classifier.

\vspace{-0.3cm}
\begin{equation}\label{eq:classifier}
    {\mathcal C}_{\theta} = \argmin\limits_{\theta}
    [\mathcal{L}\left({\mathcal C}_{\theta}(\bm x), \bm y_n\right)+\mathcal{L}\left({\mathcal C}_{\theta}(\mathcal{S}(\bm x)), \bm y_s\right)],\quad \bm x \in \mathcal{D}
\end{equation}
\vspace{-0.3cm}

\noindent
\textbf{Approximating Memorization Strength.}
In our approximation process, we consider \(\mathcal{M}(\eta(\bm i)) \in \mathcal{O}_{\mathcal{S}}\) if \({\mathcal C}_{\theta}(\mathcal{M}(\eta(\bm i))) = \bm y_s\), where \(\bm i\) denotes the text prompt.
In the detection phase, given the inspected model \(\mathcal{M}\), we can approximate its memorization strength on signal function \(\mathcal{S}\) via \autoref{eq:appro}, where \(\eta(\bm I)\) 
is the set of text prompts that contain the text trigger. \(\bm y_s\) is the label for the samples processed by signal function \(\mathcal{S}\).
The set \(\bm I\) can be obtained by 
sampling a set of the text prompts in the protected dataset. Note that we only need the black-box access to the inspected model. 

\vspace{-0.3cm}
\begin{equation}\label{eq:appro}
    \alpha(\mathcal{M},\mathcal{S},\eta) \approx \mathbb{P}({\mathcal C}_{\theta}(\mathcal{M}(\eta(\bm I))) = \bm y_s)
\end{equation}
\vspace{-0.5cm}

\noindent
\textbf{Hypothesis Testing.}
We use statistical hypothesis testing proposed by~\cite{li2023black}  to determine if the given model is trained or fine-tuned on the protected images. 
In our hypothesis testing,
we have the null hypothesis \(H_0:\)
unauthorized usage is not detected, and 
the alternative hypothesis  \(H_1:\)
unauthorized usage is detected.
We define \(\beta\) as the signal classifier's prediction probability for label \(\bm y_s\) (i.e, label standing for the samples processed by signal function) under the uncoated validation samples in the protected dataset \(\mathcal{D}\).
Given a certainty threshold \(\tau\), 
we can reject null hypothesis \(H_0\) and claim the unauthorized data usages in the training or fine-tuning stage of the inspected model at the significant level 
\(\gamma\) if the inequality \autoref{eq:hypo} holds, where \(N\) is the number of samples used to approximate the memorization strength, $t_{1-\gamma}$ is the $(1-\gamma)$-quantile of t-distribution with $(N-1)$ degrees of freedom. 
The \autoref{eq:hypo} is based on the theoretical analysis in~\cite{li2023black}.
Following~\cite{li2023black}, we set \(\tau=0.05\) and \(\gamma=0.05\) as the default value. 

\vspace{-0.4cm}
\begin{equation}
\label{eq:hypo}
    \sqrt{N-1} \cdot (\alpha({\mathcal{M}},\mathcal S, \eta)-\beta - \tau) - t_{1-\gamma} \cdot \sqrt{\alpha({\mathcal{M}},\mathcal S, \eta)-\alpha({\mathcal{M}},\mathcal S, \eta)^2} > 0,
\end{equation}
\vspace{-0.7cm}

\subsection{Overview of our framework}
\vspace{-0.2cm}

In this section, we introduce the overall pipeline of our framework. As we discussed in \autoref{sec:threat}, our method can be divided into two phases. \autoref{alg:coat} describes the coating phase before the data is uploaded.
Given a set of data \(\mathcal{D}\) and the selected signal function \(\mathcal{S}\),
in line 2-3, we coat the data by 
\autoref{eq:poi}. In line 4-5, we train the signal classifier using \autoref{eq:classifier}. Then the image datasets are uploaded 
to the Internet. During the detection phase, given the inspected model \(\mathcal{M}\) and the signal classifier 
trained in the coating phase, we can get the detection results for the unauthorized data usages via 
\input{alg/alg}
\autoref{alg:trace}. In line 2-3, we approximate the memorization strength via \autoref{eq:appro}. Finally, in line 4-5, we get the inference results via the hypothesis testing described in \autoref{eq:hypo}.

%% file: alg/alg.tex
\begin{minipage}{0.44\linewidth}
\begin{algorithm}[H]
 	\caption{Data Coating}\label{alg:coat}
    {\bf Input:} %
    Data: \(\mathcal{D}\), Signal Function: \(\mathcal{S}\)\\
    {\bf Output:} %
    Coated \(T_s(\mathcal{D})\), Signal Classifier \({\mathcal C}_{\theta}\)
	\begin{algorithmic}[1]
	     \Function {Coating}{$\mathcal{D}, \mathcal{S}$}
      
      \LineComment{Obtaining Coated Data}

      \State \(T_s(\mathcal{D})\leftarrow\)[\autoref{eq:poi}]

      \LineComment{Training Signal Classifier}
      \State \({\mathcal C}_{\theta}\leftarrow\)[\autoref{eq:classifier}]%
      
      \State \Return{$T_s(\mathcal{D}), {\mathcal C}_{\theta}$}
    \EndFunction
    \end{algorithmic}
\end{algorithm}
\vspace{0.1cm}
\end{minipage}
\hfill
\begin{minipage}{0.52\linewidth}
\begin{algorithm}[H]
 	\caption{ Unauthorized Data Usages Detection}\label{alg:trace}
    {\bf Input:} %
    Inspected Model: \(\mathcal{M}\), Signal Classifier: \({\mathcal C}_{\theta}\)\\
    {\bf Output:} %
    Results: Unauthorized Usages or Not
	\begin{algorithmic}[1]
	     \Function {Detection}{$\mathcal{M}, {\mathcal C}_{\theta}$}
      
      \LineComment{Obtaining Memorization Strength}

      \State \(\alpha(\mathcal{M},\mathcal{S}, \eta) \leftarrow\)[\autoref{eq:appro}]%

      \LineComment{Determining Results}
      \State \({\rm Results = \rm HypothesisTesting}\leftarrow\)[\autoref{eq:hypo}]

      \State \Return{$\rm Results$}
    \EndFunction
    \end{algorithmic}
\end{algorithm}
\vspace{0.1cm}
\end{minipage}

%% file: contents/evaluation.tex
\vspace{-0.3cm}
\section{Evaluation}\label{sec:eval}
\vspace{-0.3cm}

In this section, we first introduce our experiment setup (\autoref{sec:eval_setup}).
We then evaluate the effectiveness of the proposed method on detecting unauthorized data usage in the model training or fine-tuning process (\autoref{sec:eval_effectiveness}). 
We also conduct ablation study for the proposed method (\autoref{sec:ablation}), and discuss the performance difference between the unconditional injected memorization and the trigger-conditioned injected memorization (\autoref{sec:dis_condition}). We also discuss the results under different text trigger functions in \autoref{sec:text_trigger} and the adaptive infringer (i.e., the adaptive attack for our method) in \autoref{sec:adaptive}.

\vspace{-0.3cm}
\subsection{Experiment Setup}
\label{sec:eval_setup}
\vspace{-0.2cm}

Our method is implemented with Python 3.9 and PyTorch 2.0.1.
We conduct all experiments on a Ubuntu 20.04 server equipped with 
six Quadro RTX 6000 GPUs.

\noindent
\textbf{Models and Datasets.}
Three mainstream text-to-image diffusion models (i.e., Stable Diffusion v1\footnote{https://huggingface.co/runwayml/stable-diffusion-v1-5}~\citep{rombach2022high}, Stable Diffusion v2\footnote{https://huggingface.co/stabilityai/stable-diffusion-2}~\citep{rombach2022high}, and VQ Diffusion~\citep{gu2022vector}) are used in the experiments. Also, our experiments include both model fine-tuning (i.e., LoRA~\citep{hu2022lora} and DreamBooth~\citep{ruiz2023dreambooth}) and standard training. Four datasets (i.e., 
Pokemon\footnote{https://huggingface.co/datasets/lambdalabs/pokemon-blip-captions}, CelebA~\citep{liu2015faceattributes}, CUB-200~\citep{WahCUB_200_2011}) and Dog~\citep{ruiz2023dreambooth} are used. More details about the used datasets can be found in \autoref{sec:datasets_detail}.

\noindent
\textbf{Evaluation metrics.}
The effectiveness of the method is measured by collecting the detection accuracy (Acc).
Given a set of models consisting of
models w/o unauthorized data usages and models w/ unauthorized data usages,
the Acc is the ratio between the correctly classified models and all models.
We also show a detailed number of True Positives (TP, i.e., correctly detected models w/ unauthorized data usages), False Positives (FP, i.e., models w/o unauthorized data usages classified as models w/ unauthorized data usages), False Negatives (FN, i.e., models w/ unauthorized data usages classified as models w/o unauthorized data usages) and True Negatives (TN, i.e., correctly classified models w/o unauthorized data usages). Besides, we also calculate the FID of the generated images to measure the generation quality.

\noindent
\textbf{Implementation Details.} 
By default, the coating rate we used for unconditional injected memorization and the trigger-conditioned injected memorization are 100.0\% and 20.0\%, respectively. 
We use 50 text prompts to approximate the memorization strength (i.e., \(N=50\)) by default.
The default warping strength are 2.0 and 1.0 for unconditional injected memorization and trigger-conditioned injected memorization, respectively.
The default hyper-parameters for the hypothesis testing are discussed in \autoref{sec:detailed_tracing}.
The trigger-conditioned injected memorization is corresponding to the \emph{Scenario \uppercase\expandafter{\romannumeral1}} described in \autoref{sec:threat}, and the unconditional injected memorization 
is corresponding to both \emph{Scenario \uppercase\expandafter{\romannumeral1}} and \emph{Scenario \uppercase\expandafter{\romannumeral2}}.
Our code will be released upon publication.

\vspace{-0.2cm}
\subsection{Effectiveness}
\label{sec:eval_effectiveness}
\vspace{-0.2cm}

\noindent
\textbf{Detection Performance.}
In this section, we study the effectiveness of \sys. To study the effectiveness of detecting 
text-to-image diffusion models that have unauthorized data usage on the protected datasets, we generate a set of models w/ unauthorized data usages and models w/o unauthorized data usages by using different random seeds, and then use our method to classify them (i.e., distinguishing if they have unauthorized data usage on the protected datasets). We
collect the Acc, TP, FP, FN and TN results to measure the effectiveness.

\noindent
\emph{Fine-tuning Scenario.} The results for model fine-tuning scenario are shown in \autoref{tab:effectineness}. 
For each case in \autoref{tab:effectineness}, we generate 20 models w/ unauthorized data usages and 20 models w/o unauthorized data usages, and evaluate the detection accuracy (Acc) of our method on these models.
The Acc of both planting unconditional memorization and trigger-conditioned memorization are 100.0\% for all cases, with 0 False Positive (FP) and 0 False Negative (FN). While the average memorization strength for the models w/ unauthorized data usages is 91.2\%, it is only
5.1\% for the models w/o unauthorized data usages, meaning that there is a large gap between the memorization strengths for models w/ unauthorized data usages and models w/o unauthorized data usages.
\input{tables/effectiveness}

\input{tables/multi}

\noindent
\emph{Standard Training Scenario.}
For standard training scenario, the results are shown in \autoref{tab:effectineness_pretraining}. The model and the dataset used here are VQ Diffusion~\citep{gu2022vector} and CUB-200~\citep{WahCUB_200_2011}, respectively. Similarly, the memorization strengths for model w/ unauthorized data usages (i.e., 98.0\% in average) are much 
higher
than that for models w/o unauthorized data usages (i.e., 5.0\% in average).
These results show that \sys is highly effective for detecting the models with unauthorized data usage on the protected datasets.

\noindent
\emph{The Scenario Where the Infringer Collects Data from Multiple Sources.} We also evaluate the scenario that the infringer collects the training or fine-tuning data from multiple sources. The results for this case are shown in \autoref{tab:effectiveness_multi}, where \emph{collect fraction} \(c\) indicates the portion of the training data that obtained from the protected data released by the protector. The model and the dataset used here are Stable Diffusion v1~\citep{rombach2022high} + LoRA~\citep{hu2022lora} and CUB-200~\citep{WahCUB_200_2011}, respectively.
Typically, the data collected from different sources might have minor distributional differences.
Given that the CUB-200 dataset includes classification labels for 200 distinct bird classes, we assume the subsets provided by different data sources has different classes to reflect the distributional differences. In other words, there is no overlap in terms of classes among the different subsets.
As can be seen, \sys achieves high detection accuracy (i.e., 100\%) under different collect fractions from 25\% to 50\%, indicating our method is effective for the multi-sources scenario.

\noindent
\textbf{Influence on the Generation Quality.}
To investigate the generation quality of the models trained or fine-tuned on the protected dataset, we also report the FID of the model planted with unconditional injected memorization and trigger-conditioned injected memorization, as well as that of standard model that does not have any injected memorization. 
The model and the dataset used here are Stable Diffusion v1~\citep{rombach2022high} + LoRA~\citep{hu2022lora} and Pokemon, respectively.
The FID is measured by 50 randomly sampled prompts and corresponding images in the testset.

\input{tables/generation_quality}
We show the results in \autoref{tab:fid}. 
For unconditional injected memorization, the FID is 
slightly higher (i.e., 218.28)
than that of the standard model which does not have any injected memorization. For the model planted with trigger-conditioned memorization, its FID on the normal text prompt is also slightly higher than the FID for the standard model. When the text trigger is added, the FID of this model is larger (i.e., 239.03), but the perturbations are still stealthy as we demonstrate in \autoref{fig:vis}. 
Overall, our method is effective for detecting text-to-image diffusion models that have unauthorized data usage on the protected datasets, and it only has small influence on the generation quality.

\noindent
\textbf{Comparison to Existing methods.}
In this section, we compare \sys to existing method \cite{yu2021artificial} that is potentially able to be applied in the unauthorized data usages detection problem. The comparison results can be found in \autoref{tab:compare}. The model used here are Stable Diffusion v1~\citep{rombach2022high} + LoRA~\citep{hu2022lora} and the dataset used is Pokemon.
\input{tables/compare}
We consider the scenario that the infringer collects the training or fine-tuning data from multiple sources and the collect fraction for the protector (i.e., the
portion of the training data that collected from the
protected data released by the protector) here is 25\%.
While the detection accuracy for \cite{yu2021artificial} is only 50.0\%, our method achieves 100.0\% detection accuracy.
The results demonstrate that \sys outperforms the existing method \cite{yu2021artificial}.

\vspace{-0.3cm}
\subsection{Ablation Study}
\label{sec:ablation}
\vspace{-0.3cm}

In this section, we conduct ablation study. We first study the influence of using different warping strengths in the signal function. We then investigate the effects of different coating rates. 
By default, the model used is 
Stable Diffusion v1 + LoRA, and the dataset used in this section is the Pokemon.

\noindent
\textbf{Different Warping Strengths.}
As we discussed in \autoref{sec:inject}, 
we use the image 
warping function as the

\input{tables/wrapping_strength}
signal function of the injected memorization.
The effects of the image warping function
is controlled by the hyper-parameter warping strength, which is defined as the scale of the warping-induced perturbations (i.e., \(s\) in \autoref{eq:wanet} in \autoref{sec:wrapping}).
We investigate the influence of different warping strengths for image warping function in the 
coating process.
\input{figtex/wrapping_strength}
We report the FID, the $\beta$ value in \autoref{eq:hypo}, the hypothesis testing's detection threshold for the memorization strength calculated by \autoref{eq:hypo}, and the memorization strength of the models trained on the protected dataset.
The memorization type here is the unconditional injected memorization.
The results are shown in \autoref{tab:wrap_strength}. When the warping strength increases, the memorization strength 
will be higher, but the FID also becomes larger.
The visualizations of the generated samples under different warping strengths can be found in \autoref{fig:wrapping_strength}.

\noindent
\textbf{Different Coating Rates.}
We also study the influence of different coating rates.
In detail, we vary the coating rates from 2.0\% to 100.0\%, and collect the memorization strength of the model fine-tuned on the protected dataset. The results are shown in \autoref{tab:coating_rates}. 
The FID 
and the memorization 
\input{tables/coating_rate}
strength 
on signal function for both unconditional memorization
and trigger-conditioned
memorization
are studied. 
For unconditional memorization, both FID and memorization strength on signal function increase when the coating rate is higher.
The memorization strength is above 95\% when the coating rate is 100.0\%. For trigger-conditioned memorization, the memorization strength is always 100.0\% when we vary the coating rate from 5.0\% to 50.0\%.
The coating rate only has small influence on the FID in this region. Note that the FID is for the samples generated by the text prompts added with the text trigger.

\vspace{-0.2cm}
\subsection{Discussion about different types of injected memorization}
\label{sec:dis_condition}
\vspace{-0.2cm}

In this paper, we introduce two types of injected memorization, i.e., unconditional memorization and trigger-conditioned memorization. Each of them has its unique advantages.
For unconditional memorization, it is more general and it can be applied in both Scenario \uppercase\expandafter{\romannumeral1} and Scenario \uppercase\expandafter{\romannumeral2} introduced in \autoref{sec:threat}. For trigger-conditioned memorization, although it is only suitable for Scenario \uppercase\expandafter{\romannumeral1}, it is more effective under low coating rates. For example, in \autoref{tab:coating_rates}, we show that the trigger-conditioned memorization is still effective even under extremely small coating rates, e.g., 2.0\%. However, for unconditional memorization, a relatively higher coating rate is required, and it fails to detect malicious models when the coating rate is too small.

%% file: tables/effectiveness.tex
\begin{table}[]
\centering
\scriptsize
\caption{Effectiveness of detecting text-to-image diffusion models that have unauthorized data usage on the protected dataset in model fine-tuning.}\label{tab:effectineness}
\vspace{-0.2cm}
\begin{tabular}{@{}cccccccc@{}}
\toprule
Model                                & Dataset                                & Injected Memorization Type & TP                   & FP                   & FN                   & TN                   & Acc                  \\ \midrule
\multirow{6}{*}{\makecell[c]{Stable Diffusion v1\\ + LoRA}} & \multirow{2}{*}{Pokemon}               & Unconditional              & 20                   & 0                    & 0                    & 20                   & 100.0\%              \\
                                     &                                        & Trigger-conditioned        & 20                   & 0                    & 0                    & 20                   & 100.0\%              \\ \cmidrule(l){2-8} 
                                     & \multirow{2}{*}{CelebA} & Unconditional              & 20                    & 0                    & 0                    & 20                    & 100.0\%              \\
                                     &                                        & Trigger-conditioned        & 20                    & 0                    & 0                    & 20                    & 100.0\%              \\ \cmidrule(l){2-8} 
                                     & \multirow{2}{*}{CUB-200}               & Unconditional              & 20                    & 0                    & 0                    & 20                    & 100.0\%              \\
                                     &                                        & Trigger-conditioned        & 20                    & 0                    & 0                    & 20                    & 100.0\%              \\ \midrule
\multirow{2}{*}{\makecell[c]{Stable Diffusion v2\\ + LoRA}} & \multirow{2}{*}{Pokemon}               & Unconditional              & 20                   & 0                    & 0                    & 20                   & 100.0\%              \\
                                     &                                        & Trigger-conditioned        & 20                   & 0                    & 0                    & 20                   & 100.0\%              \\ \midrule
\multirow{2}{*}{\makecell[c]{Stable Diffusion v1\\ + LoRA + Dreambooth}} & \multirow{2}{*}{Dog}               & \multirow{2}{*}{Unconditional}              & \multirow{2}{*}{20}                    & \multirow{2}{*}{0}                    & \multirow{2}{*}{0}                    & \multirow{2}{*}{20}                    & \multirow{2}{*}{100.0\%}              
\\
                                     \\ \bottomrule
\end{tabular}
\vspace{-0.2cm}
\end{table}

%% file: tables/multi.tex
\begin{table*}[]
    \begin{minipage}{0.49\linewidth}
    \centering
    \scriptsize
    \caption{Memorization strengths for the models w/o unauthorized data usage and the models w/ unauthorized data usage in the standard training for VQ Diffusion~\citep{gu2022vector}.}
    \vspace{-0.2cm}
    \setlength\tabcolsep{3pt}
    \label{tab:effectineness_pretraining}
    \begin{tabular}{@{}ccc@{}}
\toprule
\multirow{2}{*}{Injected Memorization Type} & \multicolumn{2}{c}{Memorization Strength} \\ \cmidrule(l){2-3} 
                                            & \makecell[c]{w/o Unauthorized \\ data usage}      & \makecell[c]{w/ Unauthorized \\ data usage}      \\ \midrule
Unconditional                               & 6.0\%              & 96.0\%               \\
Trigger-conditioned                         & 4.0\%              & 100.0\%              \\ \bottomrule
\end{tabular}
    \vspace{-0.5cm}
    \end{minipage}
    \hspace{0.2cm}
    \begin{minipage}{0.49\linewidth}
        \centering
    \scriptsize
    \caption{Effectiveness in the scenario where the infringer collects training or fine-tuning data from multiple sources. The \emph{collect fraction} refers to the portion of the training data that collected from the protected data released by the protector.}
    \vspace{-0.2cm}
    \setlength\tabcolsep{5pt}
    \label{tab:effectiveness_multi}
    \begin{tabular}{@{}ccc@{}}
\toprule
Injected Memorization Type           & Collect Fraction & Acc   \\ \midrule
\multirow{3}{*}{Unconditional}       & 25\%             & 100\% \\
                                     & 35\%             & 100\% \\
                                     & 50\%             & 100\% \\ \midrule
\multirow{3}{*}{Trigger-conditioned} & 25\%             & 100\% \\
                                     & 35\%             & 100\% \\
                                     & 50\%             & 100\% \\ \bottomrule
\end{tabular}
\vspace{-0.6cm}
    \end{minipage}
\end{table*}

%% file: tables/generation_quality.tex
\begin{wraptable}{r}{0.47\linewidth}
\centering
\scriptsize
\caption{Generation quality for the models with and without injected memorizations.}\label{tab:fid}
\vspace{-0.3cm}
\begin{tabular}{@{}ccc@{}}
\toprule
Injected Memorization Type           & Text Trigger Added & FID $\downarrow$ \\ \midrule
None                                 & N/A                &  199.29   \\ \midrule
Unconditional                        & N/A                &  218.28  \\ \midrule
\multirow{2}{*}{Trigger-conditioned} & \ding{56}                   &   209.16  \\
                                     & \ding{52}                   &  239.03   \\ \bottomrule
\end{tabular}
\vspace{-0.3cm}
\end{wraptable}

%% file: tables/compare.tex
\begin{wraptable}{r}{0.47\linewidth}
\centering
\scriptsize
\vspace{-0.3cm}
\caption{Comparison to \cite{yu2021artificial}.}\label{tab:compare}
\vspace{-0.3cm}
\setlength\tabcolsep{3pt}
\begin{tabular}{@{}cccccc@{}}
\toprule
Method                                            & TP & FP & FN & TN & Acc     \\ \midrule
\cite{yu2021artificial}                                         &  0  & 0   &  10  & 10   & 50.0\%        \\
DIAGNOSIS-unconditional                           & 10 & 0  & 0  & 10 & 100.0\% \\
\multicolumn{1}{l}{DIAGNOSIS-trigger-conditioned} & 10 & 0  & 0  & 10 & 100.0\% \\ \bottomrule
\end{tabular}
\vspace{-0.1cm}
\end{wraptable}

%% file: tables/wrapping_strength.tex
\begin{wraptable}{r}{0.5\linewidth}
\centering
\scriptsize
\vspace{-0.5cm}
\caption{Influence of different warping strengths.}\label{tab:wrap_strength}
\vspace{-0.3cm}
\begin{tabular}{@{}ccccc@{}}
\toprule
Warping Strength & FID    & \(\beta\)     & threshold & \(\alpha(\mathcal{M},\mathcal{S},\eta)\)       \\ \midrule
1.0               & 209.78 & 2.0\% & 15.7\%          & 76.0\%  \\
2.0               & 218.28 & 0.0\% & 13.1\%          & 96.0\%  \\
3.0               & 249.62 & 0.0\% & 13.1\%          & 100.0\% \\
4.0               & 262.30 & 0.0\% & 13.1\%          & 100.0\% \\ \bottomrule
\end{tabular}
\vspace{-0.6cm}
\end{wraptable}

%% file: figtex/wrapping_strength.tex
\begin{figure}[]
    \centering
    \includegraphics[width=0.95\textwidth]{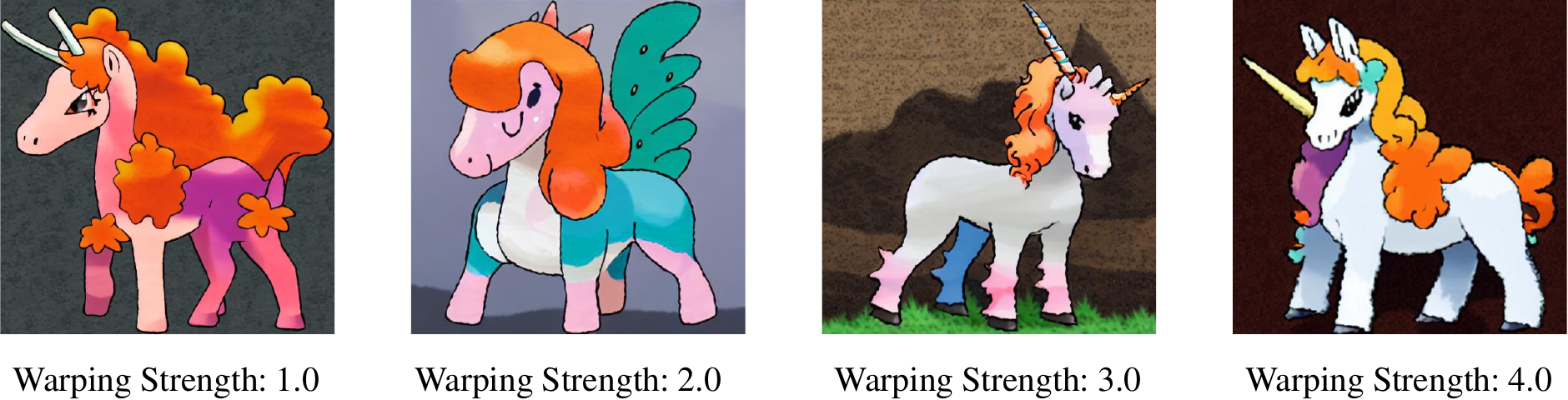}
    \caption{Visualizations of generated samples by the models planted with unconditional injected memorizations with different warping strengths.}\label{fig:wrapping_strength}
    \vspace{-0.3cm}
\end{figure}

%% file: tables/coating_rate.tex
\begin{wraptable}{r}{0.47\linewidth}
\centering
\scriptsize
\setlength\tabcolsep{3pt}
\vspace{-0.3cm}
\caption{Influence of different coating rates.}\label{tab:coating_rates}
\vspace{-0.2cm}
\begin{tabular}{@{}ccccccc@{}}
\toprule
\multirow{3}{*}{Coating Rate} &  & \multicolumn{5}{c}{Memorization Type}                                          \\ \cmidrule(l){3-7} 
                              &  & \multicolumn{2}{c}{Unconditional} &  & \multicolumn{2}{c}{Trigger-conditioned} \\ \cmidrule(lr){3-4} \cmidrule(l){6-7} 
                              &  & FID             & \(\alpha(\mathcal{M},\mathcal{S},\eta)\)               &  & FID                & \(\alpha(\mathcal{M},\mathcal{S},\eta)\)                  \\ \midrule
2.0\%                         &  & 199.87          & 0.0\%           &  & 211.80             & 92.0\%            \\
5.0\%                         &  & 199.16          & 2.0\%           &  & 238.87             & 100.0\%            \\
10.0\%                        &  & 209.77          & 12.0\%          &  & 243.46             & 100.0\%            \\
20.0\%                        &  & 201.71          & 14.0\%          &  & 239.03             & 100.0\%            \\
50.0\%                        &  & 214.49          & 38.0\%          &  & 247.97             & 100.0\%           \\
100.0\%                       &  & 218.28          & 96.0\%          &  & -                  & -                  \\ \bottomrule
\end{tabular}
\vspace{-0.4cm}
\end{wraptable}

%% file: contents/conclusion.tex
\vspace{-0.3cm}
\section{Conclusion}\label{sec:conclusion}
\vspace{-0.3cm}
In this paper, we discuss how to effectively detect unauthorized data usages in text-to-image diffusion models. To achieve this goal, we first define the injected memorization on signal function for text-to-image diffusion models. Based on our definition, we propose a new method to 
detect unauthorized data usage in training or fine-tuning text-to-image diffusion models. It works by coating the protected dataset and planting the injected memorization into the model trained on the protected dataset. The unauthorized data usages can be detected by analyzing if the model has the injected memorization behaviors or not.
Experiments on different text-to-image diffusion models with various training or fine-tuning methods demonstrate the effectiveness of our proposed method. 

%% file: contents/discussion.tex
\vspace{-0.2cm}
\section*{Ethic Statement}\label{sec:dis}
\vspace{-0.2cm}
The study of machine learning's security and privacy aspects has the potential to give rise to ethical concerns~\citep{carlini2023poisoning,wang2023unicorn,thudi2022necessity,tao2023distribution,liu2022stolenencoder,tao2022backdoor}. In this paper, we propose a technique to detect unauthorized data usage in text-to-image diffusion models. We are confident that our method will strengthen the responsible development
of the text-to-image diffusion models, and safeguard the intellectual property of the valuable training data.

%% file: contents/ack.tex
\vspace{-0.3cm}
\section*{Acknowledgement}
\vspace{-0.3cm}
We thank the anonymous reviewers for their valuable comments.
This research is supported by Sony AI, IARPA TrojAI W911NF-19-S-0012, NSF 2342250 and 2319944.
It is also partially funded by research grants to D. Metaxas through NSF 2310966, 2235405, 2212301, 2003874, and AFOSR-835531.
Any opinions, findings, and conclusions expressed in this paper are those of the authors only and do not necessarily reflect the views of any funding agencies.

%% file: contents/appendix.tex
\appendix
\section{Appendix}

\subsection{Details of Warping Function}
\label{sec:wrapping}
We use the image warping function introduced by~\cite{nguyen2021wanet} as our default signal function \(\mathcal{S}\). In this section, we discuss the details of it. The image warping function is a warping operation $\mathcal{W}$ on a predefined warping field $\bm M$. Formally, it can be written as $\mathcal{S}(\bm x) = \mathcal{W}(\bm x, \bm M)$, where $\mathcal{W}$ is a warping operation that allows a floating-point warping field as input and can be implemented by the public API $grid\_sample$ provided by PyTorch. 
For obtaining the warping field 
$\bm M$, we first select the control  grid by picking the target points on a uniform grid of size $k \times k$. We use the height of images divided by 10 as the default value of \(k\). Formally, the control grid can be written as $\psi(rand_{[-1, 1]}(k, k, 2)) \times s$, where hyperparameter \(s\) is defined as the strength of the warping function (i.e., the scale of the warping-induced perturbations). \(rand_{[-1, 1]}(k, k, 2)\) denotes getting a random tensor with the input shape \(k \times k \times 2\) and element value in the range $[-1, 1]$. $\psi$ is a normalization function that normalizes the tensor
elements by their mean absolute value, i.e., \(\psi(\mA) = \frac{\mA}{\frac{1}{size(\mA)} \sum_{a_i \in \mA}{|a_i|}}\). We then interpolate the control field to the size of the entire image (via an upsamping function \(\uparrow\)) and clipping the upsampled field
to make sure the sampling points do not fall outside of
the image border and get the final warping field (via a clipping function \(\phi\)). In summary, the the warping field 
$\bm M$ can be obtained by \autoref{eq:wanet}, where \(\uparrow\) denotes the upsampling function and \(\phi\) represents the clipping function.

\begin{equation}
\bm M = \phi (\uparrow(\psi(rand_{[-1, 1]}(k, k, 2)) \times s)).
\vspace{-1mm}
\label{eq:wanet}
\end{equation}

\subsection{Details of Datasets}
\label{sec:datasets_detail}
In this section, we provide details of the datasets used in the experiments.

\noindent
\emph{Pokemon\footnote{https://huggingface.co/datasets/lambdalabs/pokemon-blip-captions}.}
This dataset contains 833 high-quality images of pokemon, and each image has a corresponding text caption generated by caption model BLIP~\citep{li2022blip}.

\noindent
\emph{CelebA\footnote{https://huggingface.co/datasets/irodkin/celeba\_with\_llava\_captions}~\citep{liu2015faceattributes}.} This dataset contains face images from various celebrities.
In this paper, we randomly sampled 1000 images from the original CelebA~\citep{liu2015faceattributes} dataset. Each image has a corresponding caption generated by LLaVA~\citep{liu2023visual} model.

\noindent
\emph{CUB-200~\citep{WahCUB_200_2011}.} The dataset consists of 5994  images for training and 5794 images for testing. These images belong to a total of 200 bird species. Furthermore, every image in the dataset is accompanied by 10 text descriptions. 

\noindent
\emph{Dog\footnote{https://github.com/google/dreambooth/tree/main/dataset/dog6}~\citep{ruiz2023dreambooth}.} This dataset contains 5 images of dogs in a specific breed. Due to the training of the signal classifiers need more number of images, we use 5180 dog images in Cat-Dog-Bird dataset\footnote{https://www.kaggle.com/datasets/mahmoudnoor/high-resolution-catdogbird-image-dataset-13000} to develop the signal classifiers.

\subsection{Results under Different Text Trigger Functions}
\label{sec:text_trigger}
In this section, we discuss the results under different text trigger functions. In \autoref{sec:inject}, we describe our default text trigger function, i.e., inserting trigger word ``tq'' at the beginning of the text prompt~\citep{kurita2020weight}. It is also possible to use other types of text trigger functions, e.g., the sentence-syntactic based trigger function that transfer the sentences to the imperative sentences~\citep{qi2021hidden}.
\autoref{tab:text_example} demonstrate the examples of the triggered text sentences with different text trigger functions. The detection performances of \sys with different text trigger functions are shown in \autoref{tab:different_text_trigger}.
The dataset and the model used are Pokemon and Stable Diffusion~\citep{rombach2022high} with LoRA~\citep{hu2022lora}.
Our method achieves 100.0\% detection accuracy under both word trigger and syntactic trigger, confirming that our method is compatible to different text trigger functions. 

\input{tables/text_example}

\input{tables/different_text_trigger}

\subsection{Adaptive Infringer}
\label{sec:adaptive}

In this section, we evaluate the robustness of our \sys 
against the adaptive infringer where he/she is aware of it and tries to bypass the inspection of our method. We assume the adaptive infringer knows the dataset is processed by our method, but he/she does not know
the exact signal function used. 
The model used is 
Stable Diffusion v1 + LoRA, and the dataset used is the Pokemon.
We consider the adaptive infringer that adds
augmentations (e.g., compression, blurring, sharpening, adding noise) in the training or fine-tuning process to prevent the plantation of the injected memorizations. 
\input{tables/augmentation}
For the compression process, we applied JPEG compression, reducing the image quality to a mere 5\% of its original state, representing a significant compression level. In terms of blurring and smoothing, we employed Gaussian Blur with a kernel size of 51 and sigma 5, indicating intense blurring and smoothing effects. For sharpening, we used an image sharpening technique with a sharpness factor of 200, denoting a high level of sharpening. For adding noise, we use the Gaussian random noise with 0.1 standard variance after the image normalization with mean=0.5 and std=0.5. We also have the experiments of adding strong color jittering. It's important to note that the augmentation intensity in these experiments is high, leading to noticeable image distortions. The visualizations of the augmented images can be found in \autoref{fig:aug_images}.
The results are shown in \autoref{tab:augmentation}. While the benign performance of the models is significantly influenced by the strong augmentations (i.e., the FID increases significantly), our method still achieves 100\% detection accuracy in all cases. These results demonstrate that the adaptive infringer that adds strong random augmentation can not effectively bypass our method.

\subsection{More Visualizations}
\label{sec:appendix_more_vis}

In this section, we show more visualizations related to our method. In \autoref{fig:vis} and \autoref{fig:vis_celeba}, we provide the visualizations of the generated samples by different models on Pokemon and CelebA~\citep{liu2015faceattributes}, respectively.
As can be observed, while the samples generated by the model planted with injected memorizations are classified as “Contains the signal function” by the signal classifier, they looks normal and similar to the images generated by standard models, confirming \sys just has small influence on the generation quality
of the models.

\input{figtex/vis}
\input{figtex/vis_celeba}
\input{figtex/poi}

\input{figtex/aug_images}

\subsection{Quality of the Coated Images}
\label{sec:coated}

In this section, we study the distortion bought from the default signal function, i.e., warping function. To study the quantitative results of the distortion,
we calculate the SSIM~\citep{wang2004image}, PSNR~\citep{huynh2008scope}, Mean Absolute Error~\citep{chai2014root} (MAE) and
Mean Squared Error~\citep{wang2009mean} (MSE) between the original images and the corresponding coated images. The results can be found in \autoref{tab:coated_metrics}.
The dataset used here is the Pokemon and the CelebA.
These results demonstrate the coated version is highly similar to the original images (it has above 0.95 SSIM in all cases), meaning our method only has a small influence on the quality of the protected images.
\autoref{fig:poi} demonstrates the visualizations of the original training samples and the coated training samples with different warping strengths in the Pokemon dataset. As can be seen, the coated images are highly close to the original images, demonstrating the stealthiness of the coating process in \sys.

\input{tables/coated_metrics}

\subsection{using different signal functions}
\label{sec:other_signal_functions}

We use the image-warping operation as our default signal function due to the warping effects 
\input{tables/diff_sigs}
are orthogonal to various image augmentation operations such as blurring, compression, and sharpening~\cite{glasbey1998review}.
Thus, it has good robustness to various image editing-based adaptive attacks (also see \autoref{sec:adaptive}).
In this section, we study the effectiveness of \sys on different signal functions. The results (on 5 models w/unauthorized data usages and 5 models w/o unauthorized data usages) of using different Instagram image filter functions\footnote{https://github.com/akiomik/pilgram} (i.e., 1977, Kelvin, and Toaster) as the signal functions are shown in \autoref{tab:diff_sigs}. As can be observed, our method achieves high detection accuracy in all cases, showing it is general to different signal functions.
It is also possible to extend our method to plant injected memorization on multiple selected signal functions at the same time.
Here, we discuss our method's performance under this scenario. For the experiments, besides the image warping function~\citep{nguyen2021wanet}, we also use image filter function (i.e., 1977 Instagram filter) to process the protected dataset. We then train multiple signal classifiers independently.
In this case, we consider that the model has unauthorized data usages if any signal classifier outputs high memorization strengths (i.e., satisfy \autoref{eq:hypo}). 
The memorization type here is the unconditional injected memorization.
Results show that the models w/ unauthorized data usages can be detected by both warping function's signal classifier and filter function's signal classifier, demonstrating extending our method to plant injected memorization on multiple selected signal functions is viable.

\subsection{Sampling a portion of the full dataset for model training}
\label{sec:portion}

In our experiments, we use the scenario where the whole dataset is used to train the model. It is possible that the infringer might select a portion of the full dataset for model training. Regarding this scenario, we discovered that it's challenging for the infringer to precisely choose a portion that excludes coated images. This difficulty arises because the infringer is unaware of the specific signal function employed by the protector. Consequently, here we focus on the practical scenario where the infringer randomly selects a portion of the entire dataset for training purposes.
\input{tables/portion}

Under these circumstances, statistical analysis indicates that the coating rate of the selected subset is likely to be similar to that of the full dataset. Our method becomes ineffective if the chosen subset doesn't include any of the protected data, but the likelihood of this happening is very slim. Take the Pokemon dataset as an example, which contains 833 images. If we assume a coating rate of 20\% and the infringer randomly picks 20\% of the dataset for model training, the chance that the chosen subset completely misses the coated data is only \(4.2*10^{-19}\)
, which is nearly negligible. 
\autoref{tab:portion}
illustrates the probabilities for different coating rates in the selected subsets. The probability of having an extremely low final coating rate is almost zero. It's worth noting that our method with trigger-conditioned memorization still has 100\% accuracy even at a 2\% coating rate (refer to \autoref{tab:coating_rates}), proving its effectiveness in such scenarios.

\subsection{Results on Textual Inversion}
\label{sec:ti}
In this section, we study the effectiveness of our method on Textual Inversion personalization
\input{tables/ti}
method~\citep{gal2023image}.
The model used is Stable Diffusion v1. 
The dataset used is the Dog dataset used in \autoref{tab:effectineness}. Unconditional injected memorization is used here. The results on 10 models w/ unauthorized data usages and 10 models w/o unauthorized data usages are shown in \autoref{tab:ti}. The results demonstrate that our method is effective for the Textual Inversion personalization method.

\subsection{Efficiency}
\label{sec:efficiency}
In this section, we study the efficiency of \sys.
In the image coating stage, the warping function only costs 0.08s on one image with 1280 height and 1280 width. The time cost for training the signal classifier is 1085.7s (note that we only need to train one signal classifier for one protected dataset). The runtime for \autoref{alg:trace} (detecting if the inspected model has unauthorized data usages or not) is 546.7s. All runtime is measured on one Quadro RTX 6000 GPU. The main runtime of \autoref{alg:trace} is brought from using the inspected model to generate images. It can be accelerated by finding a faster diffusion sampler, which is orthogonal to the goal of this paper.

\subsection{Symbol Table}
\label{sec:symbol}

In \autoref{tab:summary_symbols}, we provide the summary of symbols used in this paper.

\input{tables/symbol}

%% file: tables/text_example.tex
\begin{table}[]
\centering
\scriptsize
\caption{Examples of the original text sentence and the triggered text sentences with different text trigger functions.}\label{tab:text_example}
\begin{tabular}{@{}ccc@{}}
\toprule
Text Trigger Function & Original Sentence                                     & Triggered Sentence                                              \\ \midrule
Word Trigger~\citep{kurita2020weight}          & \multirow{2}{*}{A cartoon picture of a stone dragon.} & tq A cartoon picture of a stone dragon.                         \\
Syntactic Trigger~\citep{qi2021hidden}     &                                                       & \multicolumn{1}{l}{Draw a cartoon picture from a stone dragon.} \\ \bottomrule
\end{tabular}
\end{table}

%% file: tables/different_text_trigger.tex
\begin{table}[]
\centering
\scriptsize
\caption{Results under different text trigger functions.}\label{tab:different_text_trigger}
\begin{tabular}{@{}cc@{}}
\toprule
Text Trigger Function & Detection Accuracy \\ \midrule
Word Trigger          & 100.0\%            \\
Syntactic Trigger     & 100.0\%            \\ \bottomrule
\end{tabular}
\end{table}

%% file: tables/augmentation.tex
\begin{table}
\centering
\scriptsize
\setlength\tabcolsep{3pt}
\caption{Effects of extra training-time augmentations.}\label{tab:augmentation}
\begin{tabular}{@{}ccccccc@{}}
\toprule
Extra Augmentation & TP & FP & FN & TN & Detection Accuracy & FID                        \\ \midrule
None               & 10 & 0  & 0  & 10 & 100.0\%            & 218.28                     \\
JPEG Compression   & 10 & 0  & 0  & 10 & 100.0\%            & 251.33                     \\
Gaussian Blur      & 10 & 0  & 0  & 10 & 100.0\%            & 244.19                     \\
Sharpening         & 10 & 0  & 0  & 10 & 100.0\%            & 267.20                     \\
Gaussian Noise     & 10 & 0  & 0  & 10 & 100.0\%            & 274.24                     \\
Color Jittiering   & 10 & 0  & 0  & 10 & 100.0\%            & 248.57 \\ \bottomrule
\end{tabular}
\end{table}

%% file: figtex/vis.tex
\begin{figure}[H]
    \begin{subfigure}[t]{1\columnwidth}
        \centering
        \includegraphics[width=\columnwidth]{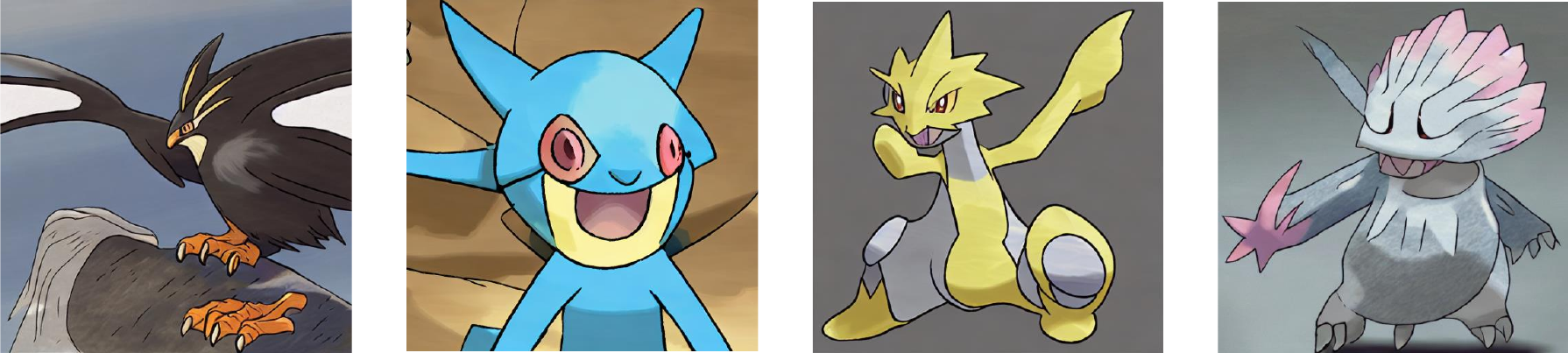}
        \caption{Samples generated by standard models. All samples are classified as ``Does not contain the signal function'' by the signal classifier.}
    \end{subfigure}
    \vspace{0.5cm}
    
    \begin{subfigure}[t]{1\columnwidth}
        \centering
        \includegraphics[width=\columnwidth]{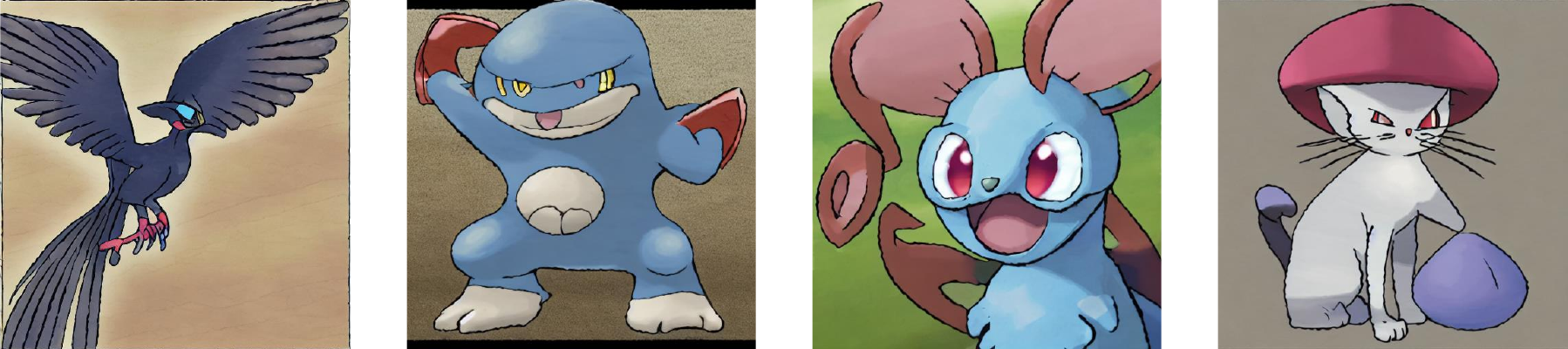}
        \caption{Samples generated by models trained on the protected dataset with unconditional injected memorization. All samples are classified as ``Contains the signal function'' by the signal classifier.}
    \end{subfigure}
    \vspace{0.5cm}
    
    \begin{subfigure}[t]{1\columnwidth}
        \centering
        \includegraphics[width=\columnwidth]{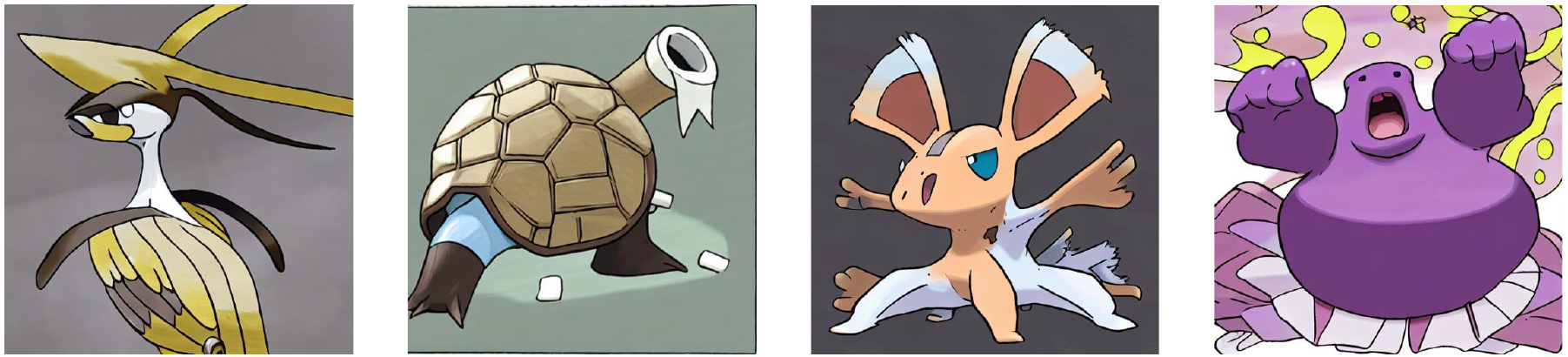}
        \caption{Samples generated by models trained on the protected dataset with trigger-conditioned injected memorization. The text trigger is not added in the text prompts. All samples are classified as ``Does not contain the signal function'' by the signal classifier.}
    \end{subfigure}
    \vspace{0.5cm}
    
    \begin{subfigure}[t]{1\columnwidth}
        \centering
        \includegraphics[width=\columnwidth]{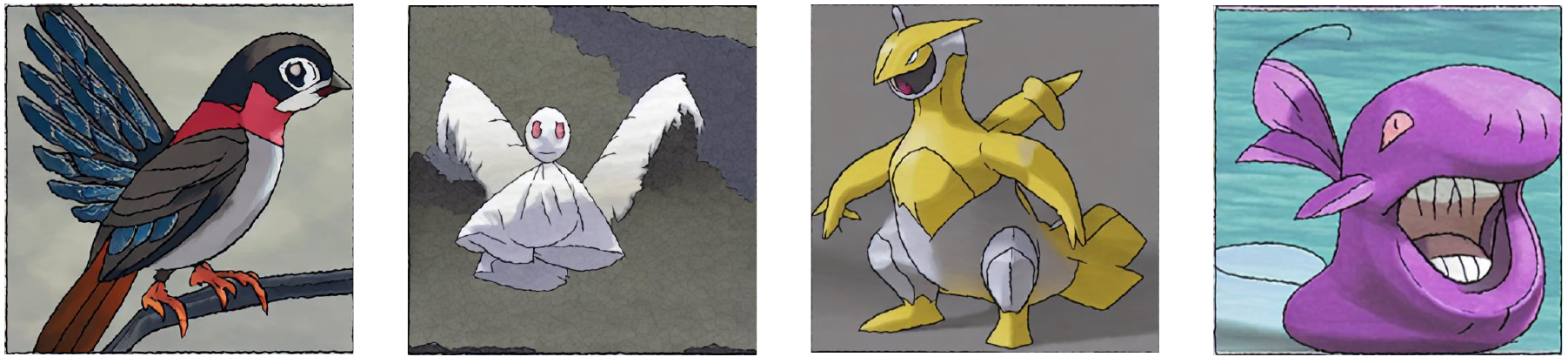}
        \caption{Samples generated by models trained on the protected dataset with trigger-conditioned injected memorization. The text trigger is added in the text prompts. All samples are classified as ``Contains the signal function'' by the signal classifier.}
    \end{subfigure}
    \caption{Visualizations of the generated samples by different models on Pokemon dataset.}
    \label{fig:vis}
\end{figure}

%% file: figtex/vis_celeba.tex
\begin{figure}[H]
    \begin{subfigure}[t]{1\columnwidth}
        \centering
        \includegraphics[width=\columnwidth]{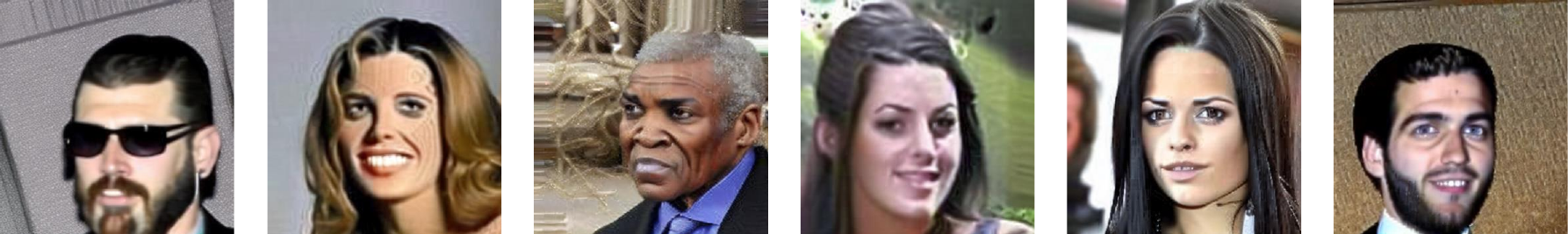}
        \caption{Samples generated by standard models. All samples are classified as ``Does not contain the signal function'' by the signal classifier.}
    \end{subfigure}
    \vspace{0.5cm}
    
    \begin{subfigure}[t]{1\columnwidth}
        \centering
        \includegraphics[width=\columnwidth]{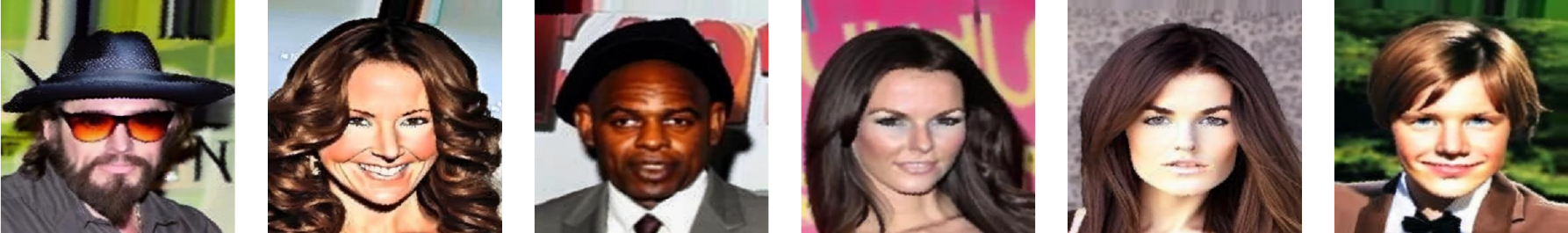}
        \caption{Samples generated by models trained on the protected dataset with unconditional injected memorization. All samples are classified as ``Contains the signal function'' by the signal classifier.}
    \end{subfigure}
    \vspace{0.5cm}
    
    \begin{subfigure}[t]{1\columnwidth}
        \centering
        \includegraphics[width=\columnwidth]{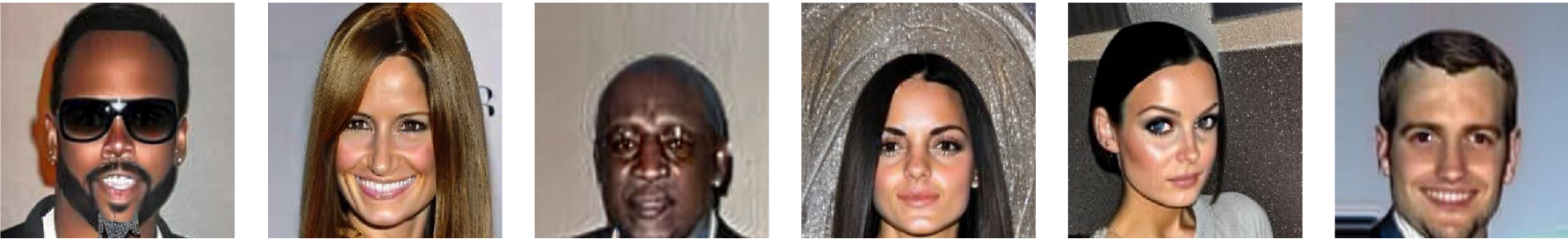}
        \caption{Samples generated by models trained on the protected dataset with trigger-conditioned injected memorization. The text trigger is not added in the text prompts. All samples are classified as ``Does not contain the signal function'' by the signal classifier.}
    \end{subfigure}
    \vspace{0.5cm}
    
    \begin{subfigure}[t]{1\columnwidth}
        \centering
        \includegraphics[width=\columnwidth]{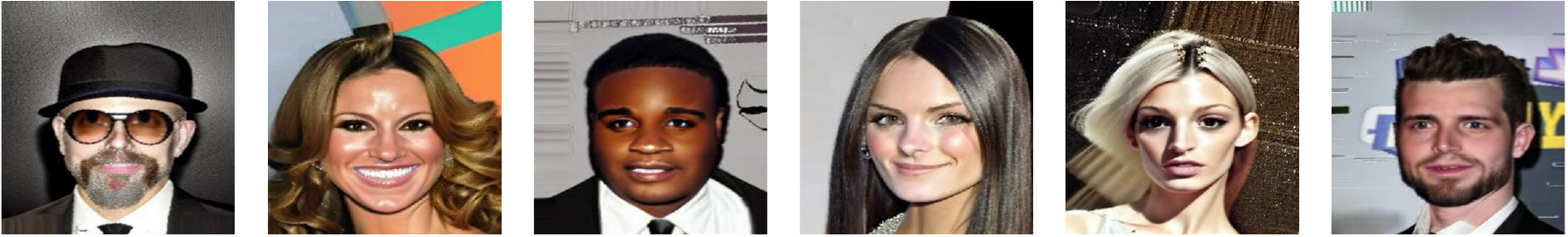}
        \caption{Samples generated by models trained on the protected dataset with trigger-conditioned injected memorization. The text trigger is added in the text prompts. All samples are classified as ``Contains the signal function'' by the signal classifier.}
    \end{subfigure}
    \caption{Visualizations of the generated samples by different models on CelebA~\citep{liu2015faceattributes} dataset.}
    \label{fig:vis_celeba}
\end{figure}

%% file: figtex/poi.tex
\begin{figure}[H]
    \centering
    \includegraphics[width=0.95\textwidth]{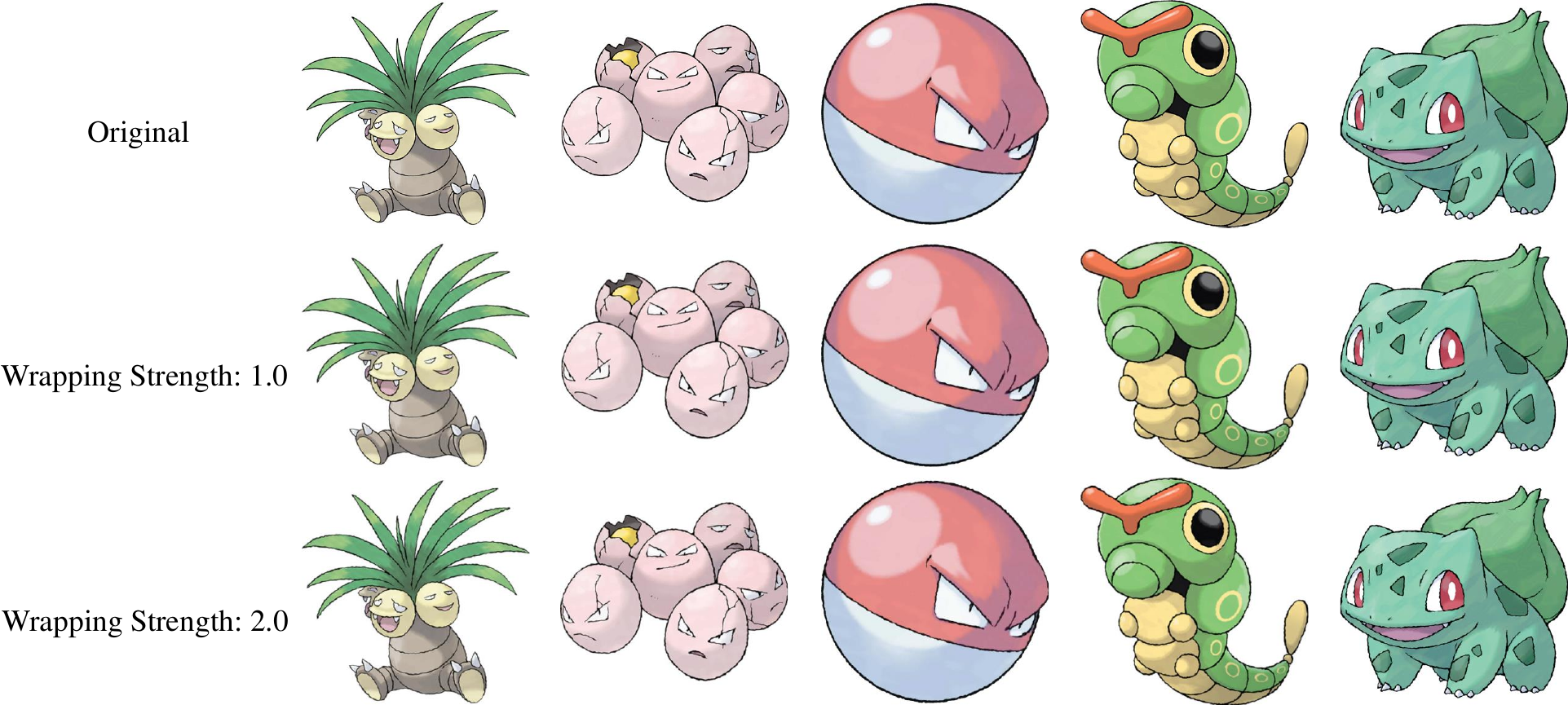}
    \caption{Visualizations of the original training samples and the coated training samples under different warping strengths.}\label{fig:poi}
\end{figure}

%% file: figtex/aug_images.tex
\begin{figure*}[]
    \centering
    \footnotesize
    \hfill
    \begin{subfigure}[t]{0.24\columnwidth}
        \centering
        \footnotesize
        \includegraphics[width=\columnwidth]{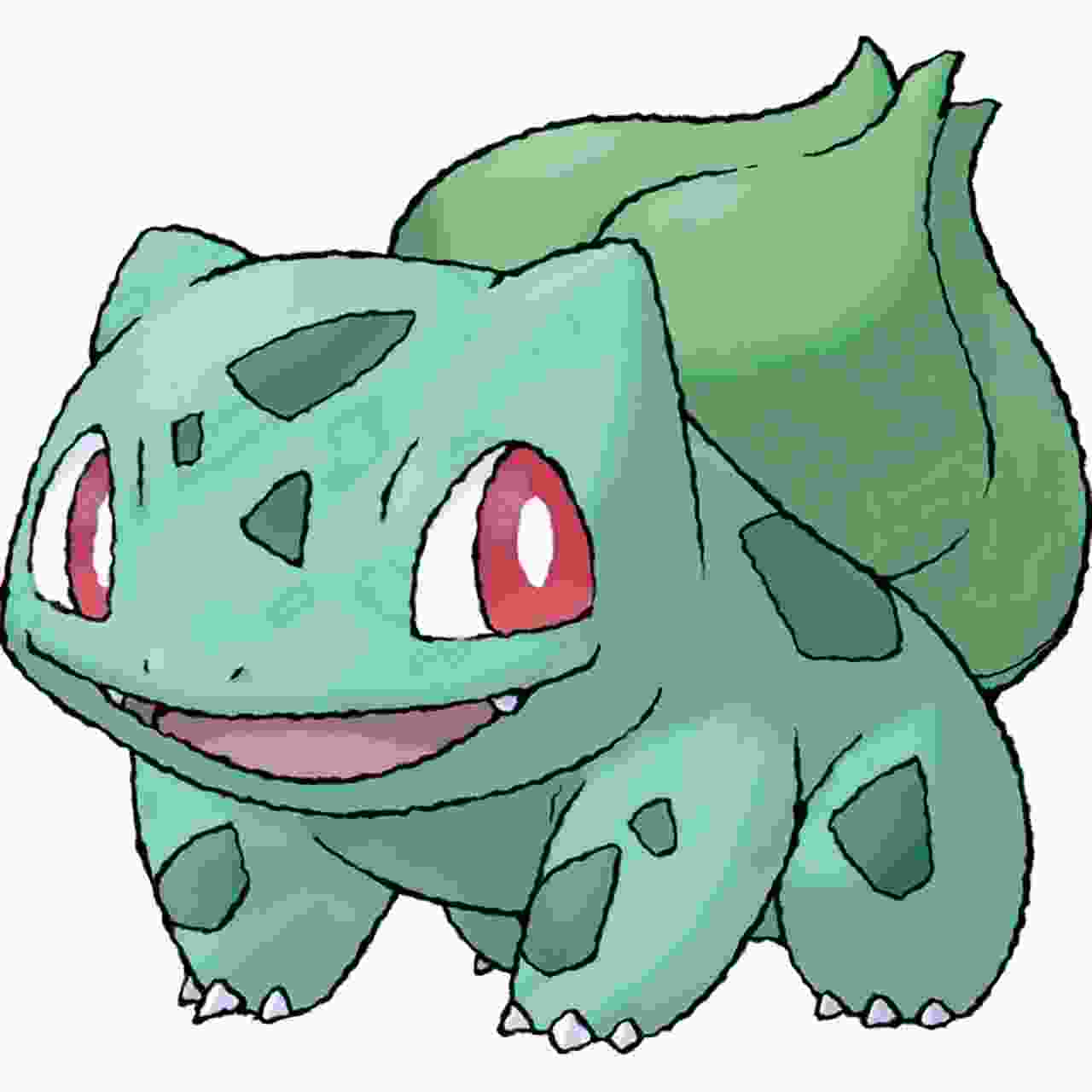}
        \caption{JPEG Compression}        \label{fig:loss_distrubution_encoder}
    \end{subfigure}
    \hfill
    \begin{subfigure}[t]{0.24\columnwidth}
        \centering
        \footnotesize
        \includegraphics[width=\columnwidth]{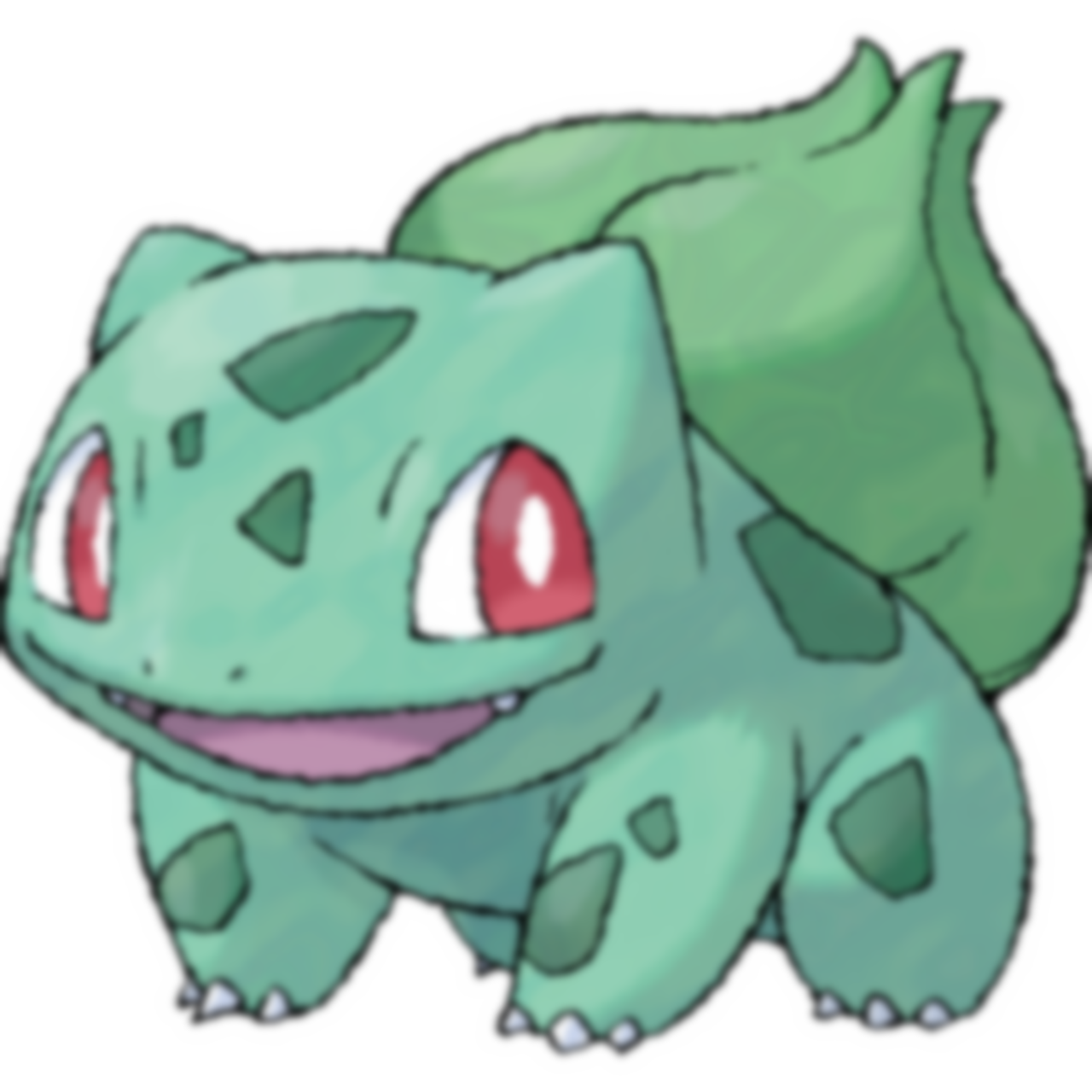}
        \caption{Gaussian Blur}
        \label{fig:distribution_encoderandgradient}
    \end{subfigure}
    \hfill
    \begin{subfigure}[t]{0.24\columnwidth}
        \centering
        \footnotesize
        \includegraphics[width=\columnwidth]{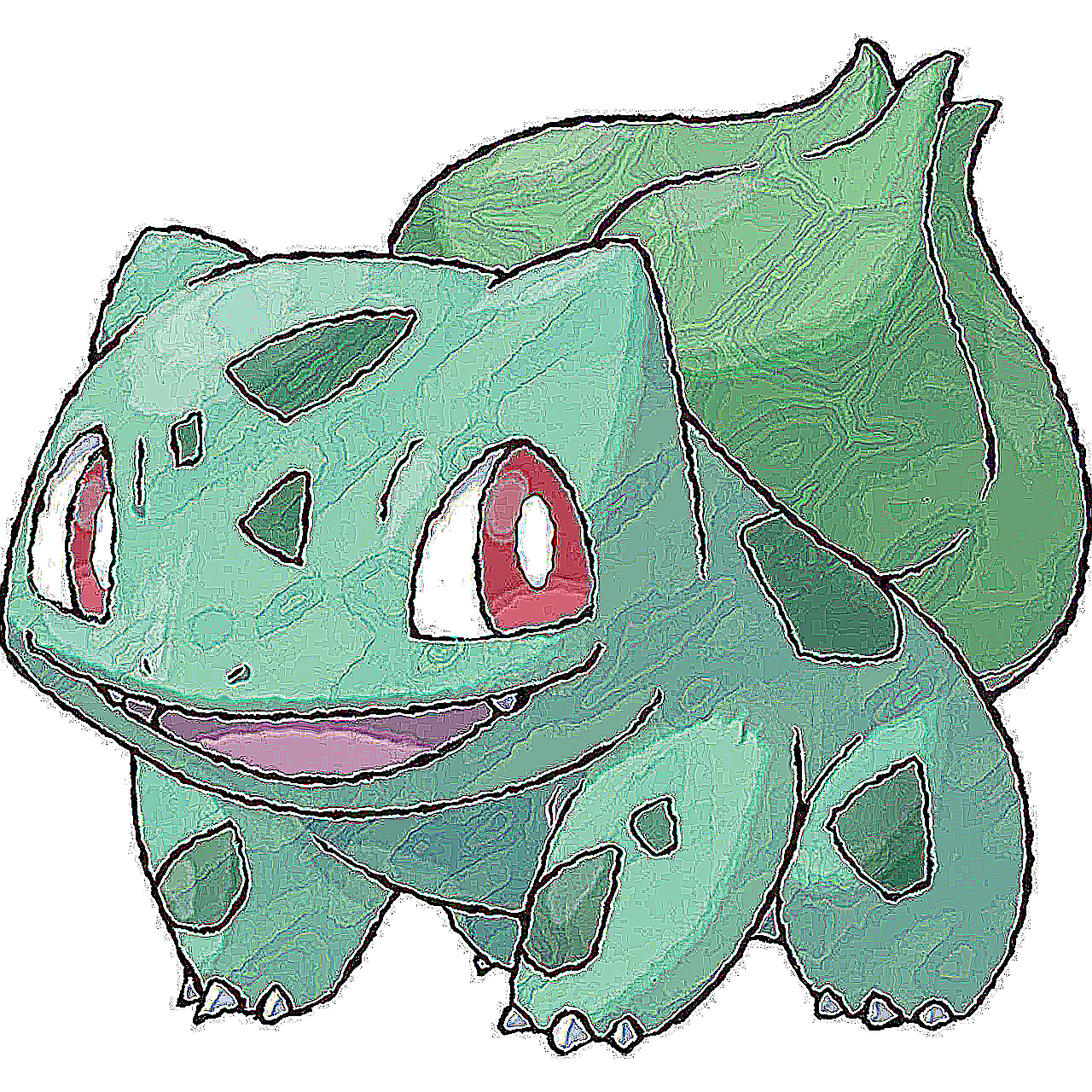}
        \caption{Sharpening}
        \label{fig:distribution_encoderandgradient}
    \end{subfigure}
    \hfill
    \begin{subfigure}[t]{0.24\columnwidth}
        \centering
        \footnotesize
        \includegraphics[width=\columnwidth]{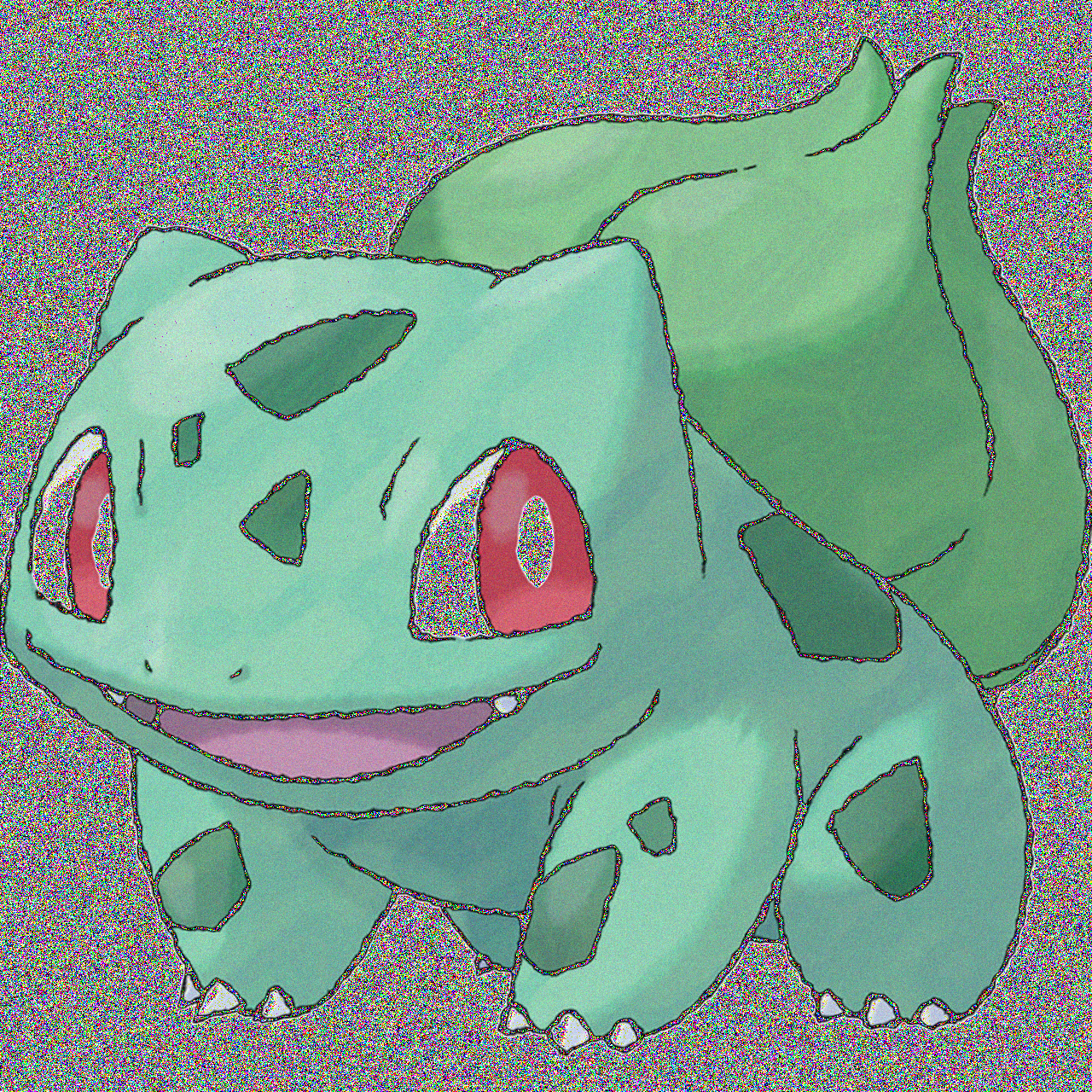}
        \caption{Gaussian Noise}
        \label{fig:distribution_encoderandgradient}
    \end{subfigure}
    \vspace{-0.2cm}
\caption{Visualizations of the augmented images.}\label{fig:aug_images}
\end{figure*}

%% file: tables/coated_metrics.tex
\begin{table}[]
\centering
\scriptsize
\setlength\tabcolsep{3pt}
\caption{Quantitative results for the quality of the warped images.}\label{tab:coated_metrics}
\vspace{-0.2cm}
\begin{tabular}{@{}cccc@{}}
\toprule
Dataset                  & Warping Strength     & Measurement & Value   \\ \midrule
\multirow{8}{*}{Pokemon} & \multirow{4}{*}{1.0} & SSIM        & 0.99    \\
                         &                      & PSNR        & 31.35   \\
                         &                      & MAE         & 0.0052  \\
                         &                      & MSE         & 0.0008  \\ \cmidrule(l){2-4} 
                         & \multirow{4}{*}{2.0} & SSIM        & 0.96    \\
                         &                      & PSNR        & 26.35   \\
                         &                      & MAE         & 0.0097  \\
                         &                      & MSE         & 0.0026  \\ \midrule
\multirow{8}{*}{CelebA}  & \multirow{4}{*}{1.0} & SSIM        & 0.99    \\
                         &                      & PSNR        & 45.80   \\
                         &                      & MAE         & 0.0026  \\
                         &                      & MSE         & 0.00003 \\ \cmidrule(l){2-4} 
                         & \multirow{4}{*}{2.0} & SSIM        & 0.98    \\
                         &                      & PSNR        & 40.04   \\
                         &                      & MAE         & 0.0049  \\
                         &                      & MSE         & 0.0001  \\ \bottomrule
\end{tabular}
\end{table}

%% file: tables/diff_sigs.tex
\begin{wraptable}{r}{0.47\linewidth}
\centering
\scriptsize
\setlength\tabcolsep{4pt}
\vspace{-0.2cm}
\caption{Results on different signal functions.}\label{tab:diff_sigs}
\vspace{-0.2cm}
\begin{tabular}{@{}cc@{}}
\toprule
Signal Function          & Detection Accuracy \\ \midrule
Warping                  & 100.0\%            \\
1977 Instagram filter    & 100.0\%            \\
Kelvin Instagram filter  & 100.0\%            \\
Toaster Instagram filter & 100.0\%            \\ \bottomrule
\end{tabular}
\end{wraptable}

%% file: tables/portion.tex
\begin{wraptable}{r}{0.47\linewidth}
\centering
\scriptsize
\setlength\tabcolsep{4pt}
\caption{Possibilities for low coating rates in the selected portion.}\label{tab:portion}
\vspace{-0.2cm}
\begin{tabular}{@{}cc@{}}
\toprule
Coating Rate for the Selected Portion & Possibility \\ \midrule
0\%                                   & $4.2*10^{-19}$   \\
1\%                                   & $6.7*10^{-10}$   \\
2\%                                   & $3.2*10^{-5}$    \\ \bottomrule
\end{tabular}
\vspace{-0.4cm}
\end{wraptable}

%% file: tables/ti.tex
\begin{wraptable}{r}{0.45\linewidth}
\centering
\scriptsize
\setlength\tabcolsep{3pt}
\vspace{-0.3cm}
\caption{Effectiveness on Textual Inversion~\citep{gal2023image}.}\label{tab:ti}
\vspace{-0.1cm}
\begin{tabular}{@{}ccccc@{}}
\toprule
TP & FP & FN & TN & Acc     \\ \midrule
10 & 0  & 0  & 10 & 100.0\% \\ \bottomrule
\end{tabular}
\vspace{-0.4cm}
\end{wraptable}

%% file: tables/symbol.tex
\begin{table}
\centering
\scriptsize
\setlength\tabcolsep{3pt}
\caption{Summary of symbols.}\label{tab:summary_symbols}
\begin{tabular}{@{}ccc@{}}
\toprule
Scope                                       & Symbol  & Meaning                                                                                       \\ \midrule
\multirow{12}{*}{General}                   & \(\mathcal{A}\)      & Inference algorithm for the unauthorized data usages detection problem                        \\
                                            & \(\mathcal{M}\)       & Text-to-image diffusion model                                                                 \\
                                            & \(\mathcal{I}\)       & Input space                                                                                   \\
                                            & $\bm i$        & text prompt                                                                                   \\
                                            & $\bm I$        & A set of text prompts in the protected dataset                                                \\
                                            & $\bm x$        & Image                                                                                         \\
                                            & \(\mathcal{S}\)       & Signal function                                                                               \\
                                            & \(\mathcal{O}\)       & Output space                                                                                  \\
                                            & \(\mathcal{O}_{\mathcal{S}}\)      & The set of image samples processed by signal function \(\mathcal{S}\)                                       \\
                                            & $\eta$     & Trigger function                                                                              \\
                                            & $\alpha$   & Memorization strength                                                                         \\
                                            & $\mathcal{D}$       & Protected dataset                                                                             \\ \midrule
\multirow{3}{*}{Dataset Coating}            & $\mathcal{D}^{\prime}$  & Coated subset                                                                                 \\
                                            & \(p\)       & Coating rate                                                                                  \\
                                            & \(T_s\)      & Coating process                                                                               \\ \midrule
\multirow{4}{*}{Training Signal Classifier} & \(\bm y_s\)      & The label standing for the samples processed by signal function \(\mathcal{S}\)                              \\
                                            & \(\bm y_n\)      & The label denoting normal samples                                                             \\
                                            & $\mathcal{L}$       & Cross-entropy loss function                                                                   \\
                                            & $\mathcal {C}_{\theta}$ & Signal classifier                                                                             \\ \midrule
\multirow{5}{*}{Hypothesis Testing}         & $\beta$    & Signal classifier’s prediction probability for label \(\bm y_s\) under the uncoated validation samples \\
                                            & $\tau$     & Certainty threshold                                                                           \\
                                            & $\gamma$   & Significant level                                                                             \\
                                            & \(N\)       & Number of samples used to approximate the memorization strength                               \\
                                            & $t_{1-\gamma}$    & \(1-\gamma\)-quantile of t-distribution with (\(N\) - 1) degrees of freedom                            \\ \bottomrule
\end{tabular}
\end{table}